\documentclass[journal]{IEEEtran}

\twocolumn
\IEEEoverridecommandlockouts
\usepackage[english]{babel}
\usepackage{amsmath,amssymb,amsfonts}
\usepackage[noend]{algpseudocode}
\usepackage{algorithm}
\usepackage{graphicx}
\usepackage{textcomp}
\usepackage{comment}
\usepackage{xcolor}
\usepackage{placeins}
\usepackage{amsmath,amsfonts,amsthm,bm}
\usepackage{multirow}
\usepackage{pdflscape}
\usepackage{tabularx}
\usepackage{lipsum}
\usepackage{subcaption}
\usepackage{csquotes}
\usepackage{url}

\usepackage{MnSymbol,bbding,pifont}
\def\BibTeX{{\rm B\kern-.05em{\sc i\kern-.025em b}\kern-.08em
    T\kern-.1
    7em\lower.7ex\hbox{E}\kern-.125emX}}
    \newcommand{\bt}[1]{\mbox{$\bf #1$}}

\usepackage{atbegshi,picture}
\usepackage{lipsum}

\AtBeginShipout{\AtBeginShipoutUpperLeft{%
  \put(\dimexpr\paperwidth -18.5cm\relax,-0.6cm){{\textcolor{red}{This paper was submitted to IEEE Transactions on Information Forensics \& Security the July 28, 2022.}}}%
}}
\begin{document}

\title{An Adaptive Method for Camera Attribution under Complex Radial Distortion Corrections \\
}
%

\author{Andrea~Montibeller,~\IEEEmembership{Student~Member,~IEEE,}
       and~Fernando~P\'erez-Gonz\'alez,~\IEEEmembership{Fellow,~IEEE}
\thanks{A. Montibeller is with the Department
of Information Engineering and Computer Science, University of Trento, Trento 38123, Italy (e-mail: andrea.montibeller@unitn.it).}
\thanks{F. P\'erez-Gonz\'alez is with the atlanTTic Research Center, Department of Signal Theory and Communications, University of Vigo, Vigo 36310, Spain (e-mail: fperez@gts.uvigo.es).}
\thanks{Manuscript received xxx xx, xxxx; revised xxx xx, xxxx.}}

\maketitle
\begin{abstract}
Radial correction distortion, applied by in-camera or out-camera software/firmware alters the supporting grid of the image so as to hamper PRNU-based camera attribution. Existing solutions to deal with this problem try to invert/estimate the correction using radial transformations parameterized with few variables in order to restrain the computational load; however, with ever more prevalent complex distortion corrections their performance is unsatisfactory. In this paper we propose an adaptive algorithm that by dividing the image into concentric annuli is able to deal with sophisticated corrections like those applied out-camera by third party software like Adobe Lightroom, Photoshop, Gimp and PT-Lens. We also introduce a statistic called cumulative peak of correlation energy (CPCE) that allows for an efficient early stopping strategy.  Experiments on a large dataset of in-camera and out-camera radially corrected images show that our solution improves the state of the art in terms of both accuracy and computational cost. 
\end{abstract}
\section{Introduction}

During the past years, camera fingerprints based on the Photo Response Non-Uniformity (PRNU) have gained broad popularity in forensic applications thanks to their ability to identify the device that captured a certain image. The PRNU is a multiplicative spatial pattern that owes its uniqueness to manufacturing imperfections that cause sensor elements to have minute area and substrate material differences that make them capture different amounts of energy even under a perfectly uniform flat field \cite{Lukas2006}. Applications of the PRNU in multimedia forensics go beyond camera identification from images \cite{Rosenfeld:2009} or videos \cite{Taspinar2020}, as they have also been used in detecting inconsistencies that reflect image manipulations \cite{Korus:2017}.     

Unfortunately, the fact that the PRNU can be accurately modeled as a white random process explains its sensitivity to geometric transformations that alter the image coordinates. Unless those transformations are reverted, standard detection statistics 
will perform poorly as they are roughly based on cross-correlations that yield very small values under grid misalignment.  In the literature several methods have been proposed to deal with those spatial transformations, including digital zoom \cite{Goljan2008Digital}, video stabilization \cite{Taspinar:2016}, high dynamic range (HDR) processing \cite{Darvish2019Camera}, and radial distortion corrections \cite{Goljian2012Sensor, Goljian2014Estimation}. It is in the context of the latter that we have developed the methodology presented in this work.  

Radial distortion correction aims at digitally removing the distortion introduced by the camera lens. This kind of processing is becoming more pervasive as devices increase their computing capabilities; in-camera correction is now common in compact models, tablets and smartphones.  On the other hand, out-camera corrections can be performed with powerful software like Adobe Lightroom, which are able to invert distortions almost perfectly by matching the model of the lens mounted on the camera. This is not done by applying conventional radial distortion models such as {\em barrel} or {\em pincushion} but by making use of complex models (i.e., with a large number of parameters).  As a consequence, existing methods \cite{Goljian2012Sensor, Goljian2014Estimation} relying on models with at most two parameters will only partially succeed in dealing with camera attribution under these complex out-camera processing. Increasing the number of model parameters often constitutes an undesirable path because reverting the distortion corrections entails a grid search whose computational load grows exponentially with the number of unknowns. 

In this work we propose a novel approach to PRNU-based camera attribution under radial corrections that is able to deal with complex models without significantly increasing the computational burden. The main idea is to divide the image under test and the PRNU into a series of concentric annuli that are thin enough to be locally describable with a simple (i.e., linear or cubic) distortion model which allows for an equally simple inverse transformation. The annuli are traversed sequentially by keeping track of the {\em cumulative peak-to-correlation energy ratio} (CPCE), which is a statistic introduced in this work and used to decide whether the radially corrected test image contains the reference PRNU. In fact, the sequential nature of the procedure makes it possible to implement an early stopping strategy to declare a match without having to process all the annuli and thus saving computational time. Another key feature of our method is {\em adaptivity}: instead of carrying out a wide-interval search for the distortion parameters describing each annulus, an adaptive Least-Mean-Squares-like predictor updates the parameters of the previously processed annulus in order to narrow down the current parameter search. This leads to a large computational efficiency without giving up flexibility. In order to steer the search we propose and justify mathematically a new objective function. 

Different variants of our method are evaluated in terms of accuracy and speed. We compare our method with the state of the art in \cite{Goljian2012Sensor} and \cite{Goljian2014Estimation} on a large dataset composed of images taken with: 1) compact devices and radially corrected in-camera, and 2) a reflex camera and radially corrected out-camera using different software tools. Our results show considerable performance improvements, especially on low-resolution images and in presence of complex radial corrections. 

The rest of the paper is organized as follows: Sect. \ref{sec:problem_form} provides the mathematical background and formulates the addressed problem. Sect. \ref{sec:prec_state_art} discusses the relevant state of the art. Sect. \ref{sec:our_method} is devoted to discussing the proposed method which in Sect. \ref{sec:experiments} is validated and compared with \cite{Goljian2012Sensor} and \cite{Goljian2014Estimation}. Finallt, Sect.\ref{sec:conclusions} presents our conclusions. 

\section{Problem Formulation and Modeling}
\label{sec:problem_form}
\subsection{Notation}
In this paper we will consider gray-scale images (the extension to color images being straightforward). Bi-dimensional signals will be denoted with boldface. For every such signal, a domain ${\mathcal S} \subset {\mathbb Z}^2$ will be specified; for instance, a signal $\bt X$ with domain ${\mathcal S}_X$ is a collection of values $X_{i,j} \in {\mathbb R}$ defined for all locations $(i,j) \in {\mathcal S}_X$. For the case of images of size $M \times N$, the original domain is  ${\mathcal I}=\{1, \cdots, M\} \times \{1, \cdots, N\} \subset {\mathbb Z}^2$; however, we will often find ourselves working with domains that are subsets of ${\mathcal I}$. We will denote by $D_2$ half of the diagonal of domain ${\mathcal I}$ measured in pixels.  Notice that the set ${\mathcal I}$ can be expressed as ${\mathcal I} = {\mathcal B} \cap {\mathbb Z}^2$, with ${\mathcal B} \subset {\mathbb R}^2$ denoting the image bounding box.   

The inner product of two signals $\bt X$ and $\bt Y$ with respective domains ${\mathcal S}_X$ and ${\mathcal S}_Y$ can be defined by extending the Frobenius product of matrices as $\langle \bt X, \bt Y \rangle \doteq \sum_{(i,j) \in {\mathcal S}} X_{i,j}Y_{i,j}$, where ${\mathcal S}={\mathcal S}_X \cap {\mathcal S}_Y$ is assumed to be non-empty. The Frobenius norm of $\bt X$ with domain ${\mathcal S}_X$ induced by this inner product is $||\bt X|| \doteq \langle \bt X, \bt X \rangle = \sum_{(i,j) \in {\mathcal S}_X} X_{i,j}^2$. The product of signals $\bt X$ and $\bt Y$, denoted by $\bt X \circ \bt Y$, is the element-wise product, i.e., $(\bt X \circ \bt Y)_{i,j} =X_{i,j} \cdot Y_{i,j}$ and is defined for all $(i,j) \in {\mathcal S}_X \cap {\mathcal S}_Y$. The multiplicative inverse of $\bt X$ is denoted by $\bt X^{\circ -1}$ and is such that $(\bt X^{\circ -1})_{i,j}=X_{i,j}^{-1}$.   
For a signal $\bt X$ with domain ${\mathcal S}_X$, we denote by $\bar{\bt X}$ a constant signal with the same support as $\bt X$ and whose value is the sample mean $\sum_{(i,j) \in {\mathcal S}_X} X_{i,j}/|{\mathcal S}_X|$, where $|{\mathcal S}_X|$ denotes the cardinality of ${\mathcal S}_X$. The normalized cross-correlation (NCC) between $\bt X$ and $\bt Y$ is defined as 
\begin{equation}
\label{eq:NCC}
 \rho(\bt X, \bt Y)= \frac{\langle \bt X - \bar{\bt X}, \bt Y -\bar{\bt Y}\rangle}{||\bt X-\bar{\bt X}||\cdot||\bt Y-\bar{\bt Y}||},
 \end{equation}
with the inner product and norms defined as above. 

Given a signal $\bt X$ with rectangular domain ${\mathcal I}$ and a vector $\bt s=(s_1,s_2) \in {\mathbb Z}^2$, we denote by $C(\bt X,\bt s)$ the cyclic shift of $\bt X$ by vector $\bt s$, so that the $(i,j)$th component of $C(\bt X,\bt s)$ is $X_{(i+s_1) \text{mod} M, (j+s_2) \text{mod} N}$. Note that the domain of  $C(\bt X,\bt s)$ is also ${\mathcal I}$. Finally, the all-zeros image is denoted by $\bt 0$.

\subsection{PRNU estimation}
As previously indicated, the PRNU is a multiplicative noise-like signal that serves as a sensor fingerprint \cite{Lukas2006},\cite{Goljan2008Digital}. Because the PRNU is a very weak signal, it is necessary to separate it from both the true image and other noise components. If  $\bt I_0$  denotes the image in absence of noise, and $\bt K$ is the PRNU, it is possible to derive the following simplified model \cite{Chen2008Determining}:
\begin{equation}
	\bt I=\bt I_{0}+\bt I_{0} \circ \bt K+\mathbf{\Theta},
	\label{eq:image}
\end{equation}
where $\mathbf{\Theta}$ is uncorrelated with both $\bt I_0$ and $\bt K$, and summarizes  noise components of different nature, and all signals are defined over ${\mathcal I}$. 
The fingerprint $\bt K$ of a camera can be extracted from $L$ images $\bt I^{(l)}$, $l=1, \cdots, L$, taken with the camera under analysis. Let $\bt W^{(l)}$ denote the noise {\em residual} obtained by applying a generic denoising filter $F(\cdot)$ to the $i$th image $\bt I^{(l)}$, as
\begin{equation}
    \bt W^{(l)} =\bt I^{(l)}-F(\bt I^{(l)}),\ \ l=1, \cdots, L.
    \label{eq:noise_residual}
\end{equation}
In all our reported experiments, we have used Mihcak's wavelet-based denoiser \cite{Mihcak1999Denoiser} for it yields an excellent trade-off between performance and complexity. Then, the PRNU can be estimated as follows \cite{Chen2008Determining}:
\begin{equation}
	{\hat {\bt K}}=\left(\sum_{l=1}^L \bt I^{(l)} \circ \bt W^{(l)} \right)\circ \left(\sum_{l=1}^L \bt I^{(l)} \circ \bt I^{(l)} \right)^{\circ -1}.
	\label{eq:camera_fingerprint}
\end{equation}
The estimate so obtained is customarily post-processed to remove some systematic artifacts that are present in most cameras. Here, we will follow \cite{Chen2008Determining} and apply a mean-removal operation by columns and rows, and a Wiener filter in the DFT aimed at removing periodic spatial artifacts. For color images, the fingerprints are estimated separately for the RGB channels and then linearly combined into gray-scale as in \cite{Goljan2009Large}.   
Given an image under investigation $\bt I$ and its corresponding residual $\bt W \doteq \bt I - F(\bt I)$ a binary hypothesis test can be formulated to decide whether $\bt I$ contains a certain PRNU $\bt K'$ for which an estimate $\hat {\bt K}'$ is available. We will denote the null hypothesis of this test (i.e., $\bt I$ does not contain ${\bt K}'$) by $H_0$ and the alternative (i.e., $\bt I$ contains ${\bt K}'$) by $H_1$.

The most popular decision statistic for the test is the  \textit{Peak-to-Correlation Energy ratio} (PCE) which computes the peak cross correlation between the test image residual $\bt W$ and the estimated PRNU $\hat{\bt K}'$ from the candidate camera, and normalizes it by an estimate of the correlation noise under $H_0$ \cite{Goljan2009Large}. 

For non-cropped images the PCE simplifies to
\begin{equation}
    \text{PCE}(\hat{\bt K}',\bt W)=\frac{\text{sgn}(\rho(\hat{\bt K}',\bt W)) \cdot \rho^2(\hat{\bt K}',\bt W)}{\frac{1}{|{\mathcal I}\backslash {\mathcal S}|}\sum_{\bt s \in {\mathcal I} \backslash {\mathcal S}} \rho^2(\hat{\bt K}',C(\bt W, \bt s))},
   \label{eq:pce}
\end{equation}
where, following the improvement proposed in \cite{Kang12}, we have included the sign of the NCC to exclude negative values that would be never expected under $H_1$. In \eqref{eq:pce}  $\mathcal S$ is a {\em cyclic exclusion neighborhood} of $(0,0)$ of small size (e.g., $11 \times 11$ pixels) to avoid contamination from cross-correlation peaks when estimating the cross-correlation noise when $H_1$ holds. Noticing that for every $\bt s$, $||C(\bt W,\bt s)-{\overline{C(\bt W,\bt s)}}|| = ||\bt W - \bar{\bt W}||$, and letting $\tilde {\bt W} \doteq \bt W - \bar{\bt W}$, \eqref{eq:pce} can be alternatively written as
\begin{equation}
    \text{PCE}(\hat{\bt K}',\bt W)=\frac{\text{ssq}(\langle \hat{\bt K}',\tilde{\bt W} \rangle)} 
    {\frac{1}{|{\mathcal I} \backslash {\mathcal S}|}\sum_{\bt s \in {\mathcal I} \backslash {\mathcal S}} \langle  \hat{\bt K}',C(\tilde{\bt W}, \bt s) \rangle^2},
   \label{eq:pce2}
\end{equation}
where we have assumed that the mean of $\hat{\bt K}'$ is zero due to the zero-meaning operation discussed above, and the signed-squared function $\text{ssq}(\cdot)$ is such that $\text{ssq}(x) \doteq \text{sgn}(x) \cdot x^2$.

\subsection{Lens Distortion Models}
\label{sec:lens_model}
To describe radially symmetric barrel/pincushion distortions we adopt the same models presented in \cite{Goljian2012Sensor} and explained in \cite{Hugemann2010Correcting, Li2005nonIterative, Janez2002Nonparametric}.
If we denote the coordinates before and after the radial distortion by $(x,y)$ and $(x',y')$, the invertible geometrical mapping $T_{\alpha}$ is given by
\begin{eqnarray}
T_{\alpha}:\mathbb{R}^2 &\rightarrow& \mathbb{R}^2 \nonumber \\
	(x,y) &\mapsto& (x',y')
	\label{eq:coordinates}
\end{eqnarray}
where
\begin{equation}
	x'=x_p+(x-x_p)(1+\alpha r^2); \\
	\label{eq:x_postRC}
\end{equation}
\begin{equation}
	y'=y_p+(y-y_p)(1+\alpha r^2), \\
	\label{eq:y_postRC}
\end{equation}
and $(x_p,y_p)$ is the {\em optical center} of the image and $r^2 \doteq [(x-x_p)^2+(y-y_p)^2]/D_2^2$ is the normalized squared radial distance from point $(x,y)$ to the optical center. This normalization by $D_2^2$ is for convenience, so that $r=1$ corresponds to half of the image diagonal \cite{Goljian2012Sensor}. Parameter
$\alpha \in \mathbb{R}$ in (\ref{eq:x_postRC}-\ref{eq:y_postRC}) models the type of radial distortion: $\alpha>0$ for pincushion distortion, and $\alpha<0$ for barrel distortion. Alternatively, given $(x_p,y_p)$ and assuming that $T_\alpha(x_p,y_p)=(x_p,y_p)$, the transformation can be written in normalized polar coordinates. Since the phase is preserved under $T_\alpha(\cdot)$, with a slight abuse of notation we will drop the phase component and sometimes write the radial transformation as $T_\alpha:\mathbb{R}^+ \cup \{0\} \rightarrow \mathbb{R}^+ \cup \{0\}$ such that
\begin{equation}
	r'=T_\alpha(r)=r(1+\alpha r^2).  \\
	\label{eq:radial_correction}
\end{equation}

More complex radial corrections  \cite{Drop2016Exact}, \cite{Goljian2014Estimation} can be expressed through an $n$th order model:
\begin{equation}
	r'=T_{\boldsymbol \alpha}(r)=r \left(1+\sum_{i=1}^n \alpha_i r^{2i}\right),  \\
	\label{eq:RC_other}
\end{equation}
where $\boldsymbol \alpha \doteq [\alpha_1, \cdots, \alpha_n]^T$ is a real parameter vector. 

Again, with some abuse of notation, and following \cite{Goljian2012Sensor}, given a signal $\bt X$ with domain ${\mathcal S}_X$, the mapping $\bt Y= T_{\boldsymbol \alpha}(\bt X)$ is produced as follows. Let $\bt X'$ be the signal with domain ${\mathcal S}_{X'}=T_{\boldsymbol \alpha}({\mathcal S}_X)$ such that, for every $(u,v) \in {\mathcal S}_X$, and with $(u',v')=T_{\boldsymbol \alpha}(u,v)$, $X'_{u',v'}=X_{u,v}$. Then, given an ouput domain ${\mathcal S}_Y$, the signal $\bt Y= T_{\boldsymbol \alpha}(\bt X)$ is obtained by interpolating the signal $\bt X'$ defined on ${\mathcal S}_X'$ at the points in ${\mathcal S}_Y$. Of course, precautions must be taken when specifying ${\mathcal S}_Y$ so that the interpolation is computable at all points in ${\mathcal S}_Y$. This aspect will be made clearer in Sect.~\ref{sec:our_method}, when we present our method.  

\subsection{Direct and inverse approaches to PCE computation}
When the image under analysis has been subjected to a radial distortion correction, the statistic $\text{PCE}(\hat{\bt K}', \bt W)$ is expected to perform poorly under $H_1$ in the hypothesis test, because the grids supporting $\hat{\bt K}'$ and $\bt W$ will not coincide (recall that the PRNU has a very narrow spatial autocorrelation function). 

The approach explored in \cite{Goljian2012Sensor} is to take into account the distortion correction when computing the PCE. If the parameter vector $\boldsymbol \alpha$ of the radial mapping is known, there are essentially two possibilities, which we will term {\em direct} and {\em inverse}. In the direct approach, the candidate PRNU $\hat{\bt K}'$ is transformed in order for its grid to match that of $\bt W$. Then, the test statistic becomes 
\begin{equation}
    	\text{PCE}_{\mathsf{dir}} ({\boldsymbol \alpha}) \doteq \text{PCE}(T_{\boldsymbol \alpha}(\hat {\bt K}'), \bt W),
	\label{eq:PCE_direct_RC}
\end{equation}
where the domain $\mathcal I_T$ of $T_{\boldsymbol \alpha}(\hat {\bt K}')$  is the largest rectangular subset of ${\mathcal I}$ for which the interpolation is computable (see discussion at the end of Sect.~\ref{sec:lens_model}) and, accordingly, ${\mathcal I}_T$ replaces ${\mathcal I}$ in the denominator of \eqref{eq:pce}.  

In the inverse approach $\bt W$ is mapped back to the original domain, so that its grid coincides with that of $\hat{\bt K}'$. Then, the test statistic in this case is  
\begin{equation}
    	\text{PCE}_{\mathsf{inv}} ({\boldsymbol \alpha}) \doteq \text{PCE}(\hat {\bt K}', T_{\boldsymbol \alpha}^{-1}(\bt W)).
	\label{eq:PCE_inverse_RC}
\end{equation}
where, as above, the domain $\mathcal I_T$ of $T_{\boldsymbol \alpha}^{-1}(\bt W)$  is the largest rectangular subset of ${\mathcal I}$ for which the interpolation is computable and ${\mathcal I}_T$ replaces ${\mathcal I}$ in the denominator of \eqref{eq:pce2}. 
  
Since one is interested in finding the best possible match,  \cite{Goljian2012Sensor} suggests using the following statistic 
\begin{equation}
		\text{PCE}_{\mathsf{max}}  ({\boldsymbol \alpha}) \doteq \max\{\text{PCE}_{\mathsf{dir}}  ({\boldsymbol \alpha}),\text{PCE}_{\mathsf{inv}}  ({\boldsymbol \alpha})\}. 
	\label{eq:PCE_max}
\end{equation}

When the parameter vector $\boldsymbol \alpha$ is not known, which is often the case in practice, it must be estimated. In \cite{Goljian2012Sensor} this is done by maximizing the test statistic in (\ref{eq:PCE_direct_RC}-\ref{eq:PCE_inverse_RC}), which makes sense from a maximum likelihood point of view. Let $\mathcal A \subset {\mathbb R}^n$ be the set of feasible vectors $\boldsymbol \alpha$; then, the statistic used in the hypothesis test is 
\begin{equation}
    	\text{PCE}_{\mathsf{max}}^* \doteq \max_{\boldsymbol \alpha \in {\mathcal A}} \text{PCE}_\mathsf{max}(\boldsymbol{\alpha}).
\end{equation}

For the case of scalar $\boldsymbol{\alpha}$ in \eqref{eq:radial_correction} the inverse radial correction $T_{\boldsymbol \alpha}^{-1}(\mathbf{W)}$ needed in \eqref{eq:PCE_inverse_RC} can be approximated via the Lagrange Inversion Theorem \cite[3.6.6.]{Abramowitz64} which yields
\begin{equation}
	r=T_{\alpha}^{-1}(r')={r'}(1-\alpha {r'}^2 + 3\alpha ^2 {r'}^{4}+ O({r'}^6)).
	\label{eq:RC_official}
\end{equation}

\begin{figure}[!ht]
		\centering
		\includegraphics[scale=0.65]{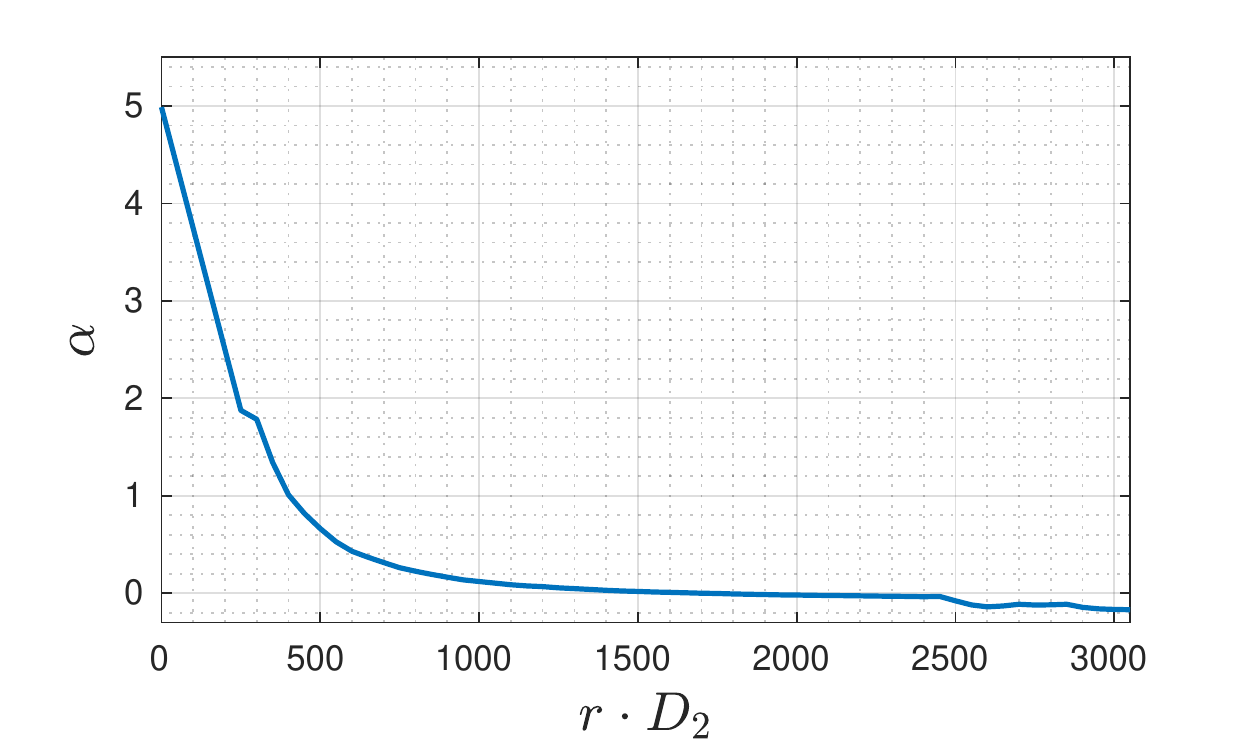}
		\caption{Values of $\alpha$ maximizing $\text{PCE}_{\mathsf{inv}}(\alpha)$ vs inner radius of the annulus. 			Values are linearly interpolated. Canon 1200D camera with EF-S 10-18mm
lens, corrected with Adobe Lightroom. Focal length: 10mm. Shutter speed:  1/100 sec. Aperture: f7.1. ISO 800. The PRNU is estimated with 20 natural images all taken with those settings.}  		
		\label{fig:alpha_trend}
\end{figure}

Using the approach described above, the radial correction can be approximately inverted in many practical cases by finding the optimal value of $\alpha$ \cite{Goljian2012Sensor}. However, when more complex radial corrections as in \eqref{eq:RC_other} have been applied, a single parameter $\alpha$ may be not sufficient. To illustrate this fact, we consider the example of an image of size $3456 \times 5184$ taken with a Canon 1200D camera using a Canon EF-S 10-18mm as lens and radially corrected with Adobe Lightroom (with settings for the mounted lens, using the strongest correction). We partitioned the image into non-overlapping annuli of width 64 pixels and found for each annulus---through exhaustive search---the value of $\alpha$ that maximizes $\text{PCE}_{\mathsf{inv}}(\alpha)$ in \eqref{eq:PCE_inverse_RC}, where inversion is done via \eqref{eq:RC_official}. The result is plotted in  Fig. \ref{fig:alpha_trend} as a function of the inner radius of the annulus. As it is quite apparent, there is a dependence of $\alpha$ with $r$ that indicates that one parameter alone is not sufficient to describe the radial transformation and that a more intricate relationship---even if parametric---must be sought. 
\section{State of the Art}
\label{sec:prec_state_art}

\begin{figure}[!ht]
		\centering
		\includegraphics[scale=0.65]{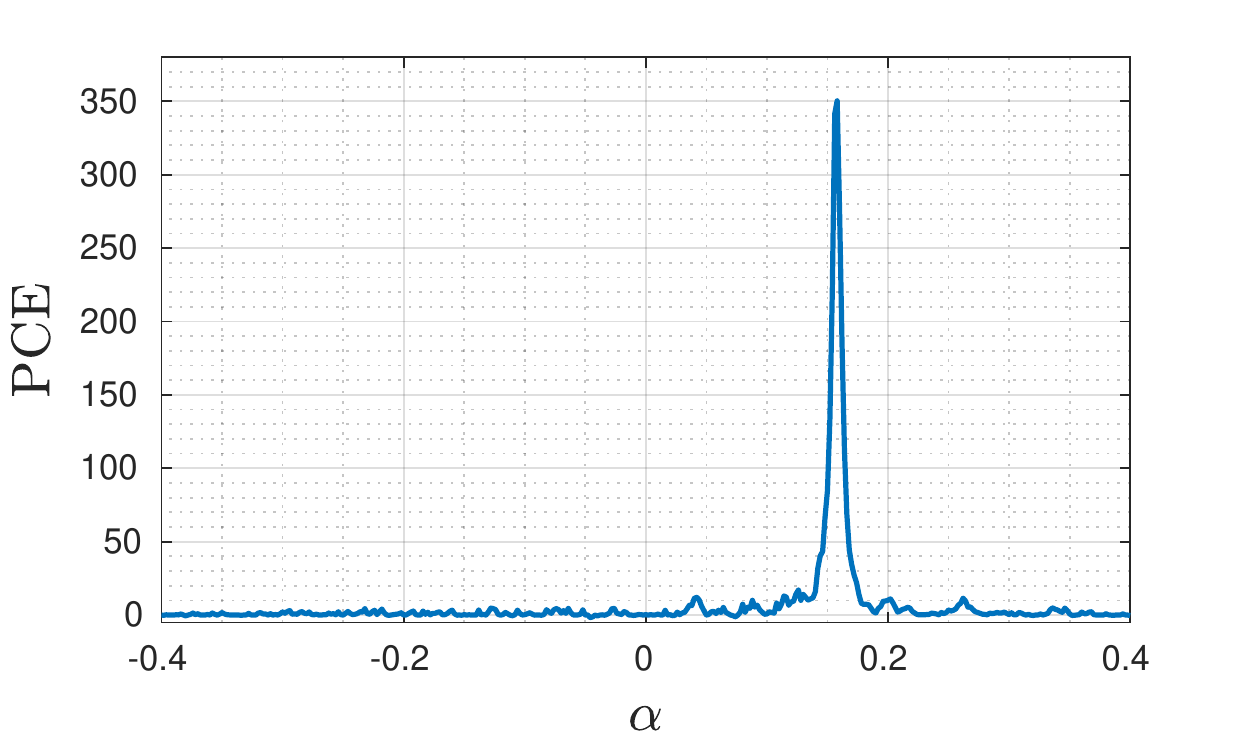}
		\caption{$\text{PCE}_{\mathsf{max}}$ as a function of $\alpha$ for a Panasonic DMC-ZS7 camera.}
		\label{fig:spiky_PCE}
\end{figure}

The PCE is very sensitive to the correct alignment of the locations corresponding to the estimated PRNU and the residual; this means that unless a value of $\boldsymbol{\alpha}$ very close to the true one is used in the mappings in \eqref{eq:PCE_direct_RC} or \eqref{eq:PCE_inverse_RC}, the resulting PCE will be very small, and hypothesis $H_1$ is likely to be rejected when it is in force.  To illustrate this phenomenon, in Fig. \ref{fig:spiky_PCE} we show the function $\text{PCE}_{\mathsf{max}}(\alpha)$ for an image taken with a Panasonic DMC-ZS7 camera, shutter speed: 1/400 s, aperture: f4.4, focal length: 19.5 mm, and ISO 100. The stepsize in $\alpha$ is $2\cdot 10^{-3}$. As we can observe, under $H_1$ the function is very spiky, with the consequence that a sufficiently dense grid must be used; otherwise, it is easy to miss the peak. In addition, this spikiness precludes the use of gradient-based algorithms, because they would only work in the very close vicinity of the peak.  

Therefore, any search grid in the parameter space has to be fine enough to be able to locate the maximum.  The method in  \cite{Goljian2012Sensor} considers that the transformations (both the direct and the inverse) are parameterized by a scalar $\alpha$ and starts by selecting a search interval $[-A,A]$ which is progressively made finer so that at each iteration $k$, with $k=1, \cdots,  k_{\mathsf{max}}$, a grid with $2^k+1$ points is generated. Note that at the $k+1$-th iteration only $2^k$ new points are produced. A threshold $\tau_1$ is set so that if, after all $k_{\mathsf{max}}$ iterations, no $\alpha$ exists in the grid such that $\text{PCE}_{\mathsf{max}} (\alpha) > \tau_1$, then the search is stopped and a mismatch is declared (i.e., $H_0$ is decided).  At every iteration, $\text{PCE}_{\mathsf{max}}$ is maximized over all grid points; this requires computing it only for the new points. Let $\alpha^\circ$ denote the grid point for which the maximum is obtained; if at some iteration $\text{PCE}_{\mathsf{max}} (\alpha^\circ) > \tau_1$, then the search stops and the algorithm proceeds to the second stage in order to refine the value of $\alpha^\circ$. 
However, in order to speed up the process, the maximization skips the exhaustive enumeration of all grid points provided that $k >4$ whenever $\alpha^\dagger$ is found such that $\text{PCE}_{\mathsf{max}} (\alpha^\dagger)>\tau_2$ (with $\tau_2>\tau_1$). In this case, the algorithm proceeds to the second stage by searching around $\alpha^\dagger$. The second stage takes the value of $\alpha$ with which the first stage was exited and constructs an interval with its two neighboring points in the grid. If $k^*$ is the exit value of $k$ for the first stage, then this interval has width $A/2^{k^*-1}$. Next, a golden section  search is performed until the width of the interval is approximately $1/(8D_2)$, with $D_2$ the half-diagonal of the image. Let $\alpha^*$ be the value found with the golden section search; then, if $\text{PCE}_{\mathsf{max}}(\alpha^*) > \tau_3$ hypothesis $H_1$ is accepted, else, $H_0$ is declared. The thresholds suggested in  \cite{Goljian2012Sensor} are $\tau_1=15$ and $\tau_2=\tau_3=75$, and $k_{\mathsf{max}}=7$. To reduce the computational load \cite{Goljian2012Sensor} downsamples the signals by a factor of two in each dimension; since this has an impact on accuracy in some cases, in the experimental section, we will consider both the downsampled (DS) and non-downsampled versions. 

The method in \cite{Goljian2014Estimation} takes a different approach to perform the inversion of radially-corrected barrel distortions by employing the so-called linear patterns that are present in the residuals and are due to artifacts of the capturing device. These patterns are typically removed towards source attribution, but when kept, they serve as pilot signals that may be used to infer the radial correction distortion. The feature that is used to steer the parameter estimation is the energy of the linear pattern, defined for a given residual $\bt W$ as $E(\bt W) \doteq ||\bt c||^2 + ||\bt r||^2$, where $\bt c$ and $\bt r$ are vectors containing respectively the column and row averages of $\bt W$. Then, considering the set of fourth-order transformations $ T_{\boldsymbol{\alpha}}(r) = r(1+\alpha_2r^2+\alpha_4r^4), $ where $\boldsymbol{\alpha} = (\alpha_2, \alpha_4)$, the method in \cite{Goljian2014Estimation} seeks to maximize $E(T_{\boldsymbol{\alpha}}^{-1}(\mathbf{W}))$ with respect to $\boldsymbol{\alpha}$, with the rationale that when the correct inverse transformation is applied, the linear pattern is recovered; otherwise, the column and row averages will be expected to produce low values. The fact that the transformation is now parameterized by two variables $\alpha_2$ and $\alpha_4$ gives more flexibility in inverting the transformation, but potentially incurs a larger computational cost. To make the optimization more manageable, a first stage consists in fitting a second-degree polynomial on variable $\alpha_2$ to values of $E(T_{\boldsymbol{\alpha}}^{-1}(\mathbf{W}))$ sampled on a grid for $\alpha_2 \in [\alpha_{\mathsf{min}}, \alpha_{\mathsf{max}}], \alpha_{\mathsf{min}} > 0$, and $\alpha_4=0$. The reason for this choice of $\alpha_4$ is that in practice the contribution of $\alpha_4$ to $T_{\boldsymbol{\alpha}}(r)$ is only significant for large $r$, that is, far from the image center. This first stage yields the value $\alpha_2^{(1)}$ of $\alpha_2$ that maximizes the difference from the energy of the linear pattern and its polynomial fit. The second stage employs a Nelder-Mead optimization (using the linear pattern energy as cost function) that is initialized with three points derived from $\alpha_2^{(1)}$. This produces the two optimal radial correction parameters $(\alpha_2^*, \alpha_4^*)$. Due to noise, the previous procedure will yield an optimum $\alpha_2 \neq 0$ regardless of whether radial correction was applied. Then, the decision is confirmed only if the cost function evaluated in a neighborhood of $(\alpha_2^*, \alpha_4^*)$ corroborates the existence of a significant peak; otherwise, the image is deemed to be not radially corrected.   

Even though, as we will see in Sect.~\ref{sec:experiments}, the performance of the two methods outlined above is rather good, they have two main intrinsic limitations that we aim at overcoming with our work: 1) their corresponding first stages employ an exhaustive search on a {\em fixed} grid. This fact, together with the high sensitivity of the PCE with respect to changes in the parameter vector $\boldsymbol{\alpha}$ about the correct one that results in a very spiky objective function, advise the use of a relatively tight grid to minimize the risk of missing the optimum. Unfortunately, this tightness entails a significant computational cost. 2) Again, due to the computational cost of an exhaustive search, the transformations $T_{\boldsymbol{\alpha}}$ and $T^{-1}_{\boldsymbol{\alpha}}$ use a small number of parameters: one in \cite{Goljian2012Sensor}, and two in \cite{Goljian2014Estimation}. Therefore, these parameterization are unable to capture more complex radial corrections, such as those employed by editing programs, a trend that is likely to increase, as the capabilities of out-of-camera processing improve. 
\section{Proposed Method}
\label{sec:our_method}

\begin{figure}[!ht]
		\centering
		\includegraphics[scale=0.55]{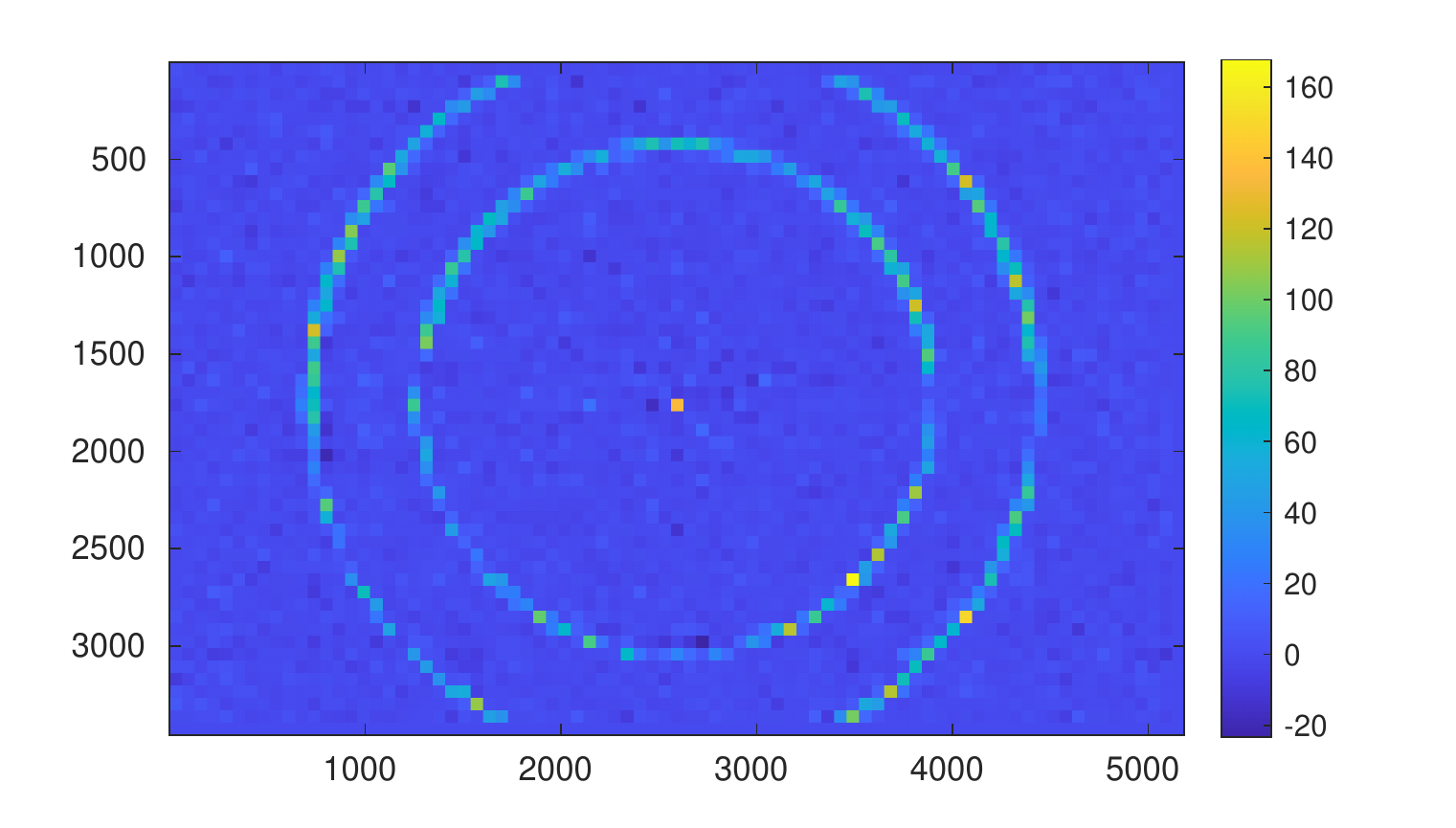}
		\caption{$\text{PCE}_\mathsf{inv}(\alpha)$ for: $\alpha=-0.01$ and $\alpha=0.05$.}
		\label{fig:PCE_map1}
\end{figure}

In order to motivate the method proposed in this paper, we will rely on an example generated with the popular photo editing software {\em Adobe Lightroom} that will give us the necessary clues. Images were taken with a Canon 1200D camera and then radially corrected with Lightroom.  In Fig.  \ref{fig:PCE_map1} we superimpose two $\text{PCE}_\mathsf{inv}$ maps (corresponding to $\alpha=-0.01$ and $\alpha=0.05$) in which $\text{PCE}_\mathsf{inv}(\alpha)$ is computed using \eqref{eq:PCE_inverse_RC} and \eqref{eq:RC_official} for non-overlapping blocks of size $64 \times 64$. For mere illustrative purposes, and in order to enhance the visibility, the (radially corrected) image under analysis (from which $\bt W$ is computed) is one of the 20 flat-field images used to extract $\hat{\bt K}'$. As we can see, the region where the PCE is significant is an annulus, and 
the position of the annulus depends on $\alpha$. This shows that if $L(r)$ denotes the radial correction induced by the software and $L^{-1}(r)$ its inverse, then for a given $\alpha=\alpha_0$, $T_{\alpha_0}^{-1}(r) \approx L^{-1}(r)$ only in a small neighborhood of some $r=r^*$.  

This experiment clearly indicates that for complex radial corrections, an approach like \eqref{eq:RC_official} will not work. However, the fact that the inversion works locally suggests breaking the problem into non-overlapping concentric annuli as shown in Fig. \ref{fig:division}, and solving each separately. 

\subsection{Set partitioning and transform computation}

\begin{figure}[!ht]
		\centering
		\includegraphics[scale=0.13]{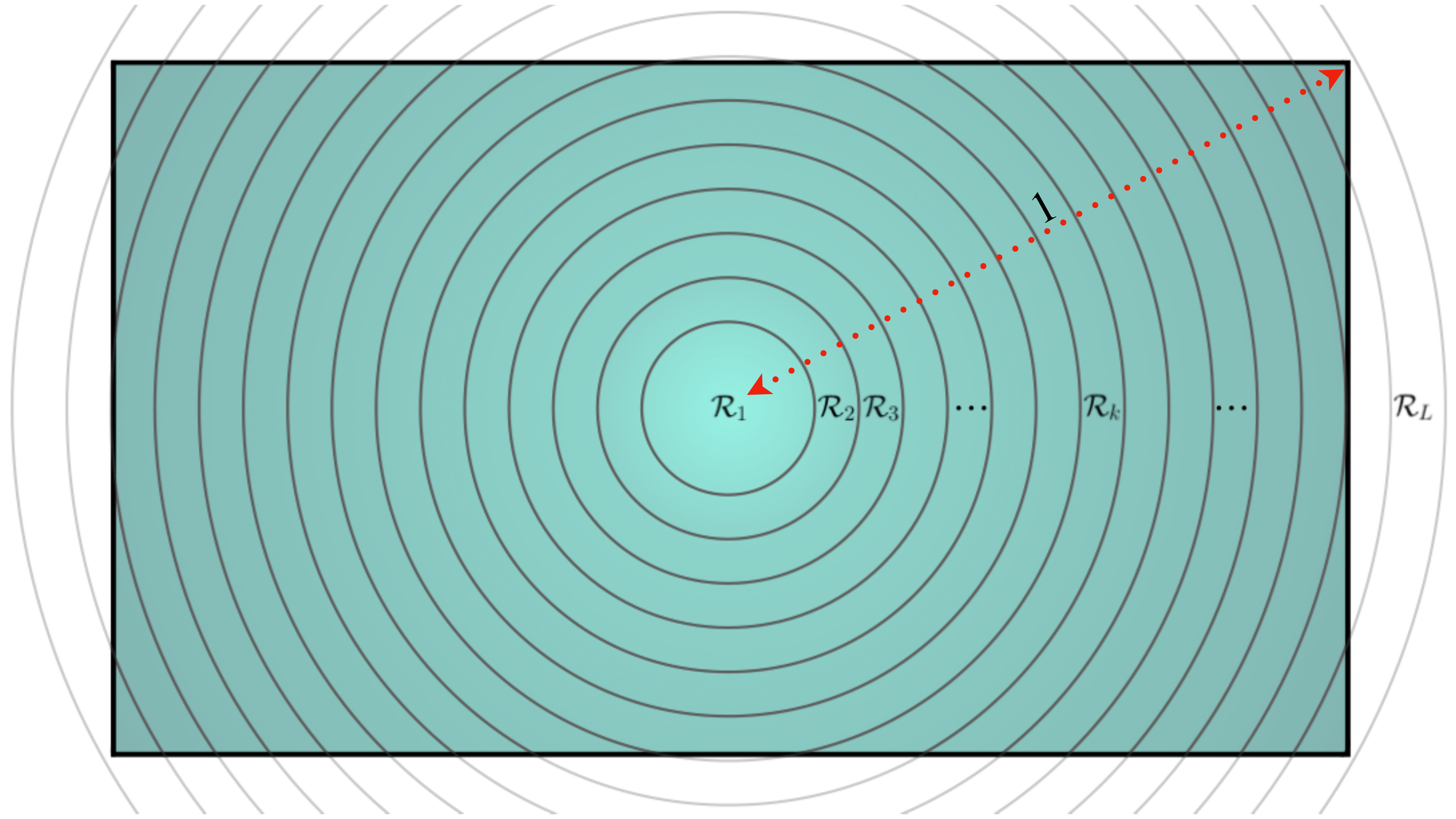}
		\caption{Annular partition used in the proposed method.}
		\label{fig:division}
\end{figure}

Let ${\mathcal R}_k$, $k=1, \cdots, L$, be the $k$th annulus described by an inner radius $r_k$ (recall that radii are scaled by $D_2$ so that $r=1$ corresponds to half of the image diagonal) and a width $\Delta_k$ as follows:
\begin{equation}
\label{eq:ring}
{\mathcal R}_k \doteq \{(u,v) \in {\mathbb R}^2: r_k^2 \leq u^2+v^2 < (r_k+\Delta_k)^2\}.
\end{equation}
The inner radii are generated as $r_{k+1}=r_k+\Delta_k$, with $r_1=0$, and the inner radius of the last annulus $r_L$ is such that $r_L < 1 <r_L+\Delta_L$ (see Fig.~\ref{fig:division}). This definition implies that the first annulus degenerates into a disk and the image is fully covered by annuli.  Except for this degenerate annulus, in this work we will assume that $\Delta_k=\Delta$ for all $k$.  

The experiment shown in Fig. \ref{fig:alpha_trend} (obtained applying a brute force search for each annulus) suggests that a good modeling of the radial correction can be obtained by allowing $\alpha$ to vary with $r$, so \eqref{eq:radial_correction} in this case becomes
\begin{equation}
\label{eq:radial_correction2}
r'=T_{\alpha(r)}(r)=r (1+\alpha(r) \cdot r^2).
\end{equation}
The idea is that by allowing $\alpha$ to be a function of $r$, we achieve much more flexibility in modeling complex distortions. Moreover, as long as the annuli are thin enough, the zero-th order approximation $\alpha(r) \approx \alpha(r_k+\Delta_k/2) \doteq \alpha_k$ will be reasonably good for all $r \in {\mathcal R}_k$. This local  approximation will allow us to use \eqref{eq:RC_official} for the inverse transform. However, since we are allowing $\alpha$ to vary with $r$, instead of a locally cubic dependence, as in \eqref{eq:radial_correction2}, it also makes sense to consider a locally linear one, i.e., $r'=T_{\alpha(r)}(r)=r (1+\alpha(r))$. Even though for the generic mappings we will keep using $T_{\alpha_k}(r)$ and $T^{-1}_{\alpha_k}(r)$ for the sake of generality, we specialize them by adding the sub-indices $c$ to denote cubic, and $l$ to denote linear. Therefore, on each annulus we write
\begin{eqnarray}
\label{eq:direct2}
T_{\alpha_k, c} (r) &\doteq& r (1+\alpha_k \cdot r^2); \ \ \nonumber \\
T_{\alpha_k, l} (r) &\doteq& r (1+\alpha_k), \ \ r \in {\mathcal R}_k,
\end{eqnarray}
whereas the corresponding inverse mappings are
\begin{eqnarray}
\label{eq:inverse2}
T_{\alpha_k, c}^{-1} (r') &\approx& r' (1- \alpha_k \cdot r'^2+3 \alpha_k^2 r'^4), \ \ r' \in T_{\alpha_k, c}({\mathcal R}_k); \ \ \nonumber\\
T_{\alpha_k, l} ^{-1}(r') &=& \frac{r'}{1+\alpha_k}, \ \ r' \in T_{\alpha_k, l}({\mathcal R}_k),
\end{eqnarray}
Note that the ranges of the inverse transforms in \eqref{eq:inverse2} may be different because the image of each annulus will differ under the locally cubic and locally linear mappings. 

Given a collection of annuli ${\mathcal R}_k$, $k=1, \cdots, L$, one can see the mapping $T_{\boldsymbol \alpha}(r)$ in \eqref{eq:PCE_direct_RC} as a sequence of transformations  $T_{\alpha_k}(r)$, $k=1, \cdots, L$, that is parameterized by a vector $\boldsymbol \alpha=[\alpha_1, \cdots, \alpha_L]^T$. Obviously, the maximization of the PCE with respect to $\boldsymbol \alpha \in {\mathcal A} \doteq {\mathcal A}_1 \times \cdots {\mathcal A}_L$, with ${\mathcal A_k}$ the feasible set for $\alpha_k$,  would suffer from a combinatorial explosion due to the $L$ dimensions involved, so we will be interested in finding efficient alternative ways for performing an approximate maximization.

A first step is to treat each annulus separately and find the optimal value of $\alpha_k$ constrained to the $k$th annulus. There are several possible approaches at this stage. One would be to find  $\alpha_k$ that maximizes the PCE constrained to the  $k$th annulus; unfortunately, since the total PCE {\em is not} the sum of those constrained PCEs, it is quite difficult to work individually with each annulus using such a criterion. Instead, we have opted for a maximum likelihood estimation approach that aims at finding the $\alpha_k$ that has the highest likelihood of producing the observed cross-correlations with the estimated PRNU. Once we describe how the optimal $\alpha_k$ can be found for each annulus in an adaptive way (Sect.~\ref{sec:adaptive}), we proceed by explaining how the PCE can be computed and updated (Sect.~\ref{sec:PCE_comp}).  

In the following, we give a formal description of the annuli for the inverse approach (i.e., using $T^{-1}_{\alpha_k}$) and afterwards indicate how to adapt the discussion to the direct approach. Let ${\mathcal P}_k$ be the set of points of the image grid that are contained in the $k$th annulus, i.e., 
\begin{equation}
\label{eq:intersect}
{\mathcal P}_k \doteq \left( D_2 \cdot {\mathcal R}_k \right) \cap \mathcal I, \ \ k=1, \cdots, L,
\end{equation}
where multiplication of ${\mathcal R}_k$ by $D_2$ (i.e., half the diagonal in pixels) is necessary to re-scale the annulus back to integer-valued coordinates (recall that $r=1$ corresponds to half the diagonal). 

\begin{figure}[!ht]
		\centering
		\includegraphics[width=\linewidth]{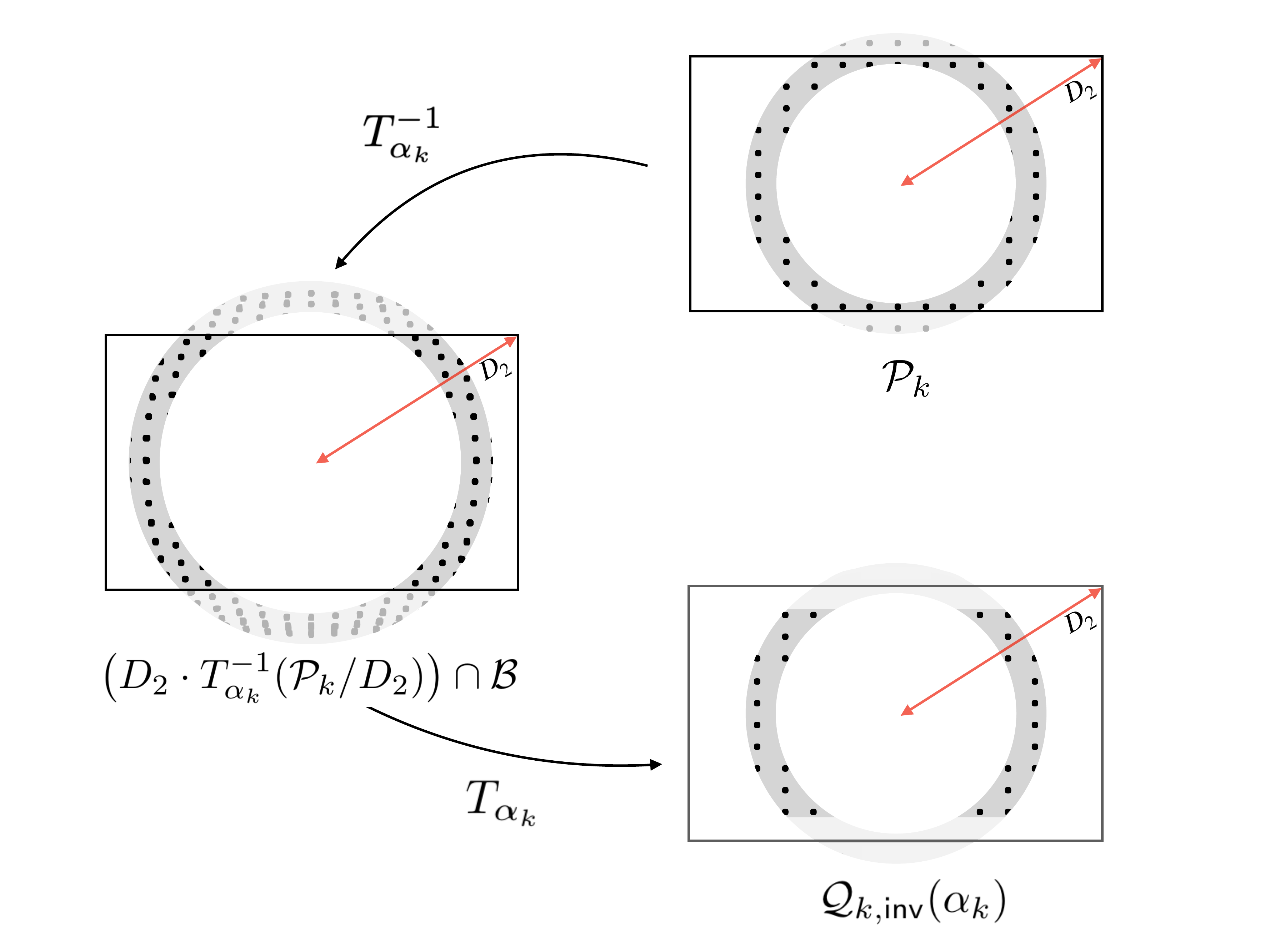}
		\caption{Illustration of the application of transforms $T^{-1}_{\alpha_k}$ and $T_{\alpha_k}$, and related domains.}
		\label{fig:Transforms}
\end{figure}

Given $\tilde{\bt W}=\bt W-\bar{\bt W}$ and ${\mathcal P}_k$, computation of $T_{\alpha_k}^{-1}(\tilde{\bt W})$ proceeds as follows (see Fig.~\ref{fig:Transforms}). First, the image of the set ${\mathcal P}_k$ under $T^{-1}_{\alpha_k}$ , i.e. $T^{-1}_{\alpha_k}({\mathcal P}_k)$ is calculated and the transformed points lying outside the image boundaries ${\mathcal B}$ are discarded, as the subsequent interpolation would not be computable. For the remaining points, $T_{\alpha_k}^{-1}(\tilde{\bt W})$ is obtained by interpolation from $\tilde{\bt W}$. We let ${\mathcal Q}_{k,\mathsf{inv}}(\alpha_k)$ be the set of points of ${\mathcal P}_k$ for which their image under $T^{-1}_{\alpha_k}$ exists (the sub-index $\mathsf{inv}$ stands for `inverse approach'). Formally, this set is
\begin{equation}
{\mathcal Q}_{k,\mathsf{inv}}(\alpha_k) =  D_2\cdot T_{\alpha_k}\left( \left[\left(D_2\cdot T^{-1}_{\alpha_k}({\mathcal P_k}/D_2) \right) \cap {\mathcal B}\right]/D_2\right). 
\end{equation}

Notice that if the set ${\mathcal P}_k$  transformed via $T_{\alpha_k}^{-1}$  does not get out of the image bounds ${\mathcal B}$, then ${\mathcal Q}_{k,\mathsf{inv}}(\alpha_k)={\mathcal P_k}$; otherwise, ${\mathcal Q}_{k,\mathsf{inv}}(\alpha_k) \subset {\mathcal P_k}$. As a consequence, ${\mathcal Q}_{k,\mathsf{inv}}(\alpha_k) \cap {\mathcal P}_k = {\mathcal Q}_{k,\mathsf{inv}}(\alpha_k)$.
Also notice that, as explicitly indicated, the set ${\mathcal Q}_{k,\mathsf{inv}}(\alpha_k)$ and, in particular, its cardinality, varies with $\alpha_k$. 

For the direct approach, the considerations are similar. Basically, we have to exchange the roles of $T_{\alpha_k}$ and $T_{\alpha_k}^{-1}$. Recalling that the sub-index $\mathsf{dir}$ stands for `direct approach', the set ${\mathcal Q}_{k,\mathsf{dir}}(\alpha_k)$ can be formally written as
\begin{equation}
{\mathcal Q}_{k,\mathsf{dir}}(\alpha_k) =  D_2\cdot T^{-1}_{\alpha_k}\left( \left[\left(D_2\cdot T_{\alpha_k}({\mathcal P_k}/D_2) \right)  \cap {\mathcal B}\right]/D_2\right). 
\end{equation}

\subsection{Optimization with respect to $\alpha_k$}
\label{sec:optimization}

Once the annuli have been characterized, in this section we address the problem of finding the optimal values of $\alpha_k$ that parameterize the transformations $T_{\alpha_k}^{-1}$ and $T_{\alpha_k}$ for the $k$th annulus. 

For the sake of compactness, we will find it useful to denote the cross-correlation and the energy of the transformed residual computed over ${\mathcal Q}_{k,\mathsf{inv}}(\alpha_k)$ as, respectively, 
\begin{eqnarray}
\label{eq:Phi}
\mathsf{\Phi}_{k,\mathsf{inv}}(\alpha_k) &\doteq& \sum_{(i,j) \in  {\mathcal Q}_{k,\mathsf{inv}}(\alpha_k)} \hat K'_{i,j} \cdot \left[T^{-1}_{\alpha_k} (\tilde{\bt W})\right]_{i,j}, \\
\label{eq:E}
\mathsf{E}_{k,\mathsf{inv}}(\alpha_k) &\doteq& \sum_{(i,j) \in {\mathcal Q}_{k,\mathsf{inv}}(\alpha_k)} \left[T^{-1}_{\alpha_k} (\tilde{\bt W})\right]_{i,j}^2,
\end{eqnarray}
by making implicit the use of the inverse transformation $T^{-1}_{\alpha_k}(\cdot)$, and $\hat{\bt K}'$ and $\tilde{\bt W}$.  Similarly, we denote by $\mathsf{\Phi}_{k,\mathsf{dir}}(\alpha_k)$ and $\mathsf{E}_{k,\mathsf{dir}}(\alpha_k)$ the cross-correlation and energy for the direct mapping $T_{\alpha_k}(\cdot)$ computed over ${\mathcal Q}_{k,\mathsf{dir}}(\alpha_k)$.

In Appendix \ref{app:estimator_alpha} we derive an estimator of $\alpha_k$ on the $k$th annulus. This estimator is rooted in the principle of maximum likelihood applied to the output of a bank of cross-correlations. For the inverse approach, this becomes 
\begin{equation}
\label{eq:argmax_i}
\alpha^*_k =\arg \max_{\alpha_k \in {\mathcal A}_k} \varphi_{k,\mathsf{inv}}(\alpha_k),
\end{equation}
where
\begin{equation}
\label{eq:varphi_inv}
\varphi_{k,\mathsf{inv}}(\alpha_k) \doteq \frac{\mathsf{\Phi}_{k,\mathsf{inv}}(\alpha_k)}{\mathsf{E}_{k,\mathsf{inv}}(\alpha_k)}.
\end{equation}

For the direct approach, the optimization is carried out after replacing the subindex $\mathsf{inv}$ by $\mathsf{dir}$ in both \eqref{eq:argmax_i} and \eqref{eq:varphi_inv}. We notice the proposed objective function is different from the PCE (constrained to the $k$th annulus); besides the theoretical justification in Appendix \ref{app:estimator_alpha}, in \cite{tech-report:Montibeller-Perez} we provide empirical evidence that optimization of our objective function renders better global performance. 

\subsection{Adaptive optimization}
\label{sec:adaptive}

One key observation from Fig.~\ref{fig:alpha_trend} is that the sequence $\alpha_k^*$, $k=1, \cdots, L$, changes smoothly for sufficiently small $\Delta_k$. This hints at the possibility of reducing the computational complexity of the exhaustive search by using an adaptive predictor. In our case, we will show experimentally that a linear predictor $\bt u$ with length $U$ suffices to achieve excellent results. In the following, we explain this adaptive procedure. As above, we will give the details for the inverse approach, as the direct one is methodologically identical. 

First, we need to select an initial index that we will denote by $k_0$. To this end, we  look for the annulus that gives the best results under no transformations (i.e., when $\alpha_k=0$). Formally, this implies that
\begin{equation}
k_0 = \arg \max_{k=1, \cdots, L} \varphi_{k,\mathsf{inv}}(0).
\end{equation} 
Once this initial point is found, the optimal value of $\alpha_{k_0}$ is found by exhaustive search in a discrete set around $\alpha_{k_0}=0$. Let ${\mathcal A}_{k_0}$ be such a neighborhood, then following \eqref{eq:argmax_i}, $\alpha_{k_0}^*= \arg \max_{\alpha_{k_0} \in {\mathcal A}_{k_0}} \varphi_{k_0,\mathsf{inv}}(\alpha_{k_0})$. 

We will find it useful to define an auxiliary sequence $\{\beta_k\}$ that is initialized as $\beta_k=\alpha^*_{k_0} \cdot \delta_{k-k_0}$, where $\delta_k$ is Kronecker's delta.\footnote{Although from a notational point of view, it would be more correct to define a sequence for every iteration of the algorithm, we allow replacing values in this sequence in order to avoid overcomplicating the notation.} This sequence is used to store the regressor values.  Since the starting point is $k=k_0$, there are two possible directions for the prediction: forward (i.e., $k > k_0$), and backward (i.e, $k < k_0$).\footnote{Degenerate cases arise when $k_0=L$ or $k_0=1$, for which the forward and backward predictions, respectively, are not needed.} We will describe how the former is carried out, and then indicate the modifications needed for the latter. We define the forward regressor at index $k$ as $\boldsymbol{\beta}_k^T \doteq [\beta_{k-U+1}, \cdots,  \beta_{k-1}, \beta_{k}]$, where $U$ is the length. Notice that as a consequence of initializing the auxiliary sequence, $\boldsymbol{\beta}_{k_0}^T = [0, \cdots, 0, \alpha^*_{k_0}]$. We also need a vector of weights at index $k$ that will be denoted by $\bt u_{k}$; this vector of length $U$ is initialized as $\bt u_{k_0}^T=[0, \cdots, 0, 1]$. Then, for $k > k_0$ the output of the predictor at index $k$ will be computed as 
\begin{equation}
\label{eq:predictor}
\hat \alpha_k = \bt u_{k-1}^T \boldsymbol \beta_{k-1}, 
\end{equation}
for $k=k_0+1, \cdots, L$. This predicted value is refined by exhaustive search in a discrete neighborhood of $\hat \alpha_k$. Let ${\mathcal A}_k$ denote such a neighborhood; then $\alpha_k^*$ is obtained as in \eqref{eq:argmax_i}. The details on how the neighborhood ${\mathcal A}_k$ is constructed are given below. Before that, we explain the updating procedure for $\bt u_k$ and $\boldsymbol \beta_k$. To that end, we define the {\em a posteriori error} at index $k$ as
\begin{equation}
\label{eq:error}
e_k \doteq \alpha_k^*-\hat \alpha_k,
\end{equation}
for  $k=k_0+1, \cdots, L$.
This error is used to drive the adaptive algorithm. It is easy to show that the gradient vector of $|e_k|^2$ with respect to the weights vector $\bt u_{k-1}$ is equal to $-2 e_k \boldsymbol  \beta_{k-1}$. Then, following the Least Mean Squares algorithm \cite{Widrow85}, we propose to update the weights by taking a step in the direction of the negative gradient, that is, 
\begin{equation}
\bt u_{k}=\bt u_{k-1} + \mu  e_k \boldsymbol  \beta_{k-1}, \ \ k=k_0+1, \cdots, L,
\end{equation}
where $\mu$ is the so-called step-size. The update of the sequence $\{\beta_k\}$ containing the regressor is done by making $\beta_{k}=\alpha_k^*$; the forward regressor vector $\boldsymbol \beta_k$ is updated accordingly. This iterative procedure is then repeated by going back to \eqref{eq:predictor} and proceeding until the sequence $\alpha_{k_0+1}^*, \alpha_{k_0+2}^*, \cdots, \alpha_L^*$ is produced.    
 
 The backward prediction proceeds in a similar way, but now vector $\boldsymbol{\beta}_k$ is defined as $\boldsymbol{\beta}_k^T \doteq [\beta_k, \beta_{k+1}, \cdots, \beta_{k+U-1}]$; this means that at the backward initialization, vector $\boldsymbol \beta_{k_0}$ will take advantage of the availability of values of $\alpha_k^*$ that have been already computed, i.e., $\boldsymbol \beta_{k_0}=[\alpha^*_{k_0}, \alpha^*_{k_0+1}, \cdots, \alpha^*_{k_0+U-1}]^T$. The weights vector for the backward prediction $\bt u_{k_0}$ is initialized as $\bt u_{k_0}=[1, 0, \cdots, 0]$.  Now this weights vector is updated in the reverse direction:
\begin{equation}
\bt u_{k}=\bt u_{k+1} + \mu  e_k \boldsymbol  \beta_{k+1}, \ \ k=k_0-1, \cdots, 1,
\end{equation} 
and again the sequence $\{\beta_k\}$ containing the regressor is updated by making $\beta_{k}=\alpha_k^*$; the backward regressor vector $\boldsymbol \beta_k$ is updated accordingly. The algorithm thus generates the sequence $\alpha_{k_0-1}^*, \alpha_{k_0-2}^*, \cdots, \alpha_1^*$. 

After both forward and backward predictions are finished, the optimal vector is $\boldsymbol \alpha^*_\mathsf{inv} = [\alpha_1^*, \cdots, \alpha_L^*]^T \in {\mathcal A}$, where once again we have added the subindex $\mathsf{inv}$ to stress the fact that we are dealing with the inverse approach. The same procedure applied to the direct approach will yield an optimal vector $\boldsymbol \alpha^*_\mathsf{dir}$. The pseudo-code for the proposed algorithm is provided in the technical report \cite{tech-report:Montibeller-Perez}.\footnote{The code is available at
\url{https://github.com/AMontiB/AdaptivePRNUCameraAttribution}}

One critical point of the algorithm is the refining of $\hat \alpha_k$ that produces $\alpha_k^*$. While smarter strategies might be possible, here we perform an exhaustive search around $\hat \alpha_k$ in a discrete set ${\mathcal A}_k$. Of course, the cardinality of this set must be kept at a small value in order to limit the computational burden. On the other hand, the discrete points must be generated finely enough to output a value that is sufficiently close to the optimal. We thus employ two parameters to describe the set: $\lambda_k$ that controls the resolution, and $A_k$ that is an odd integer that determines the number of points. Then, given $\hat \alpha_k$ and these parameters, the search set is constructed as:
\begin{equation}
\label{eq:A_k}
{\mathcal A}_k = \{\hat \alpha_k + \lambda_k \cdot n: n \in {\mathbb Z} \cap [-(A_k-1)/2,(A_k-1)/2]\}.
\end{equation}
Note that this construction guarantees that $|{\mathcal A}_k|=A_k$. The parameter $\lambda_k$ is selected to be commensurate with $|\alpha^*_k-\alpha^*_{k-1}|$ in the forward case (resp. $|\alpha^*_k-\alpha^*_{k+1}|$ in the backward case), so that the smaller the change in $\alpha_k^*$, the finer the grid. In Sect.~\ref{sec:default} we give more details about the rules that were employed to generate $\lambda_k$ for the experiments. Regarding the size of the set $A_k$, this is updated in the same loop as the predictor; for the forward predictor, the rule is as follows: if for index $k$ the maximum $\alpha_{k}^*$ is found at one of the extremes of the set ${\mathcal A}_{k}$ (i.e., $\alpha^*_{k}=\hat \alpha_{k} - \lambda_{k} \cdot (A_{k}-1)/2$ or $\alpha^*_{k}=\hat \alpha_{k} + \lambda_{k} \cdot (A_{k}-1)/2$) then the size of the set is increased at the following iteration, i.e., $A_{k+1}=A_{k}+2$. Otherwise, if $A_{k}$ is already small, i.e., $A_{k}=A_{\text{min}}$ for some minimum size $A_{\text{min}}$, then $A_{k+1}=A_{\text{min}}$; else (i.e, if the maximum in ${\mathcal A}_{k}$ is not found at either of the extremes, and the set is large enough), the size is decreased at the following iteration, i.e., $A_{k+1}=A_{k}-2$. This update is intended to find a compromise between the size of the set and the objective of capturing the optimal $\alpha_k$. For the backward prediction the reasoning is identical, but updating $A_{k-1}$ from $A_k$ (see \cite{tech-report:Montibeller-Perez} for an example of the evolution of $A_k$).

\subsection{PCE computation for the optimal $\boldsymbol \alpha$}
\label{sec:PCE_comp}

As a result of the adaptive algorithm presented in the previous section, it is possible to compute the PCEs that are required in the hypothesis test, that is, $\text{PCE}_\mathsf{inv}(\boldsymbol \alpha^*_\mathsf{inv})$ and $\text{PCE}_\mathsf{dir}(\boldsymbol \alpha^*_\mathsf{dir})$, see the definitions in \eqref{eq:PCE_direct_RC} and \eqref{eq:PCE_inverse_RC}. In both cases, the numerator and denominator of the PCE are already available, as they are required for the optimization. The only additional computations are simple sums to accumulate the results corresponding to the different annuli. To see how this is so for the inverse approach, notice first that the right hand side of \eqref{eq:PCE_inverse_RC} requires computing the difference $T^{-1}_{\boldsymbol \alpha^*}(\bt W)-\overline{T^{-1}_{\boldsymbol \alpha^*}(\bt W)}$ (cf. the expression of the PCE in \eqref{eq:pce2}), which can be simplified by noticing that:  1) It is reasonable to write $\overline{T^{-1}_{\boldsymbol \alpha^*}(\bt W)} \approx T^{-1}_{\boldsymbol \alpha^*}(\bar{\bt W})$ because $T^{-1}_{\boldsymbol \alpha^*}$ is a geometrical transformation that will not substantially alter the mean value of the residual.\footnote{Strict equality does not hold because $\bar{\bt W}$ and $T^{-1}_{\boldsymbol \alpha^*} (\bar{\bt W})$ do not have the same support.} 2) Due to zero-meaning on the residual, it is possible to write $\bar{\bt W}=\bt 0$. With these considerations,  we can write $T^{-1}_{\boldsymbol \alpha^*}(\bt W)-\overline{T^{-1}_{\boldsymbol \alpha^*}(\bt W)} \approx T^{-1}_{\boldsymbol \alpha^*}(\tilde{\bt W})$, which is simpler to compute.

With this approximation, the numerator of  $\text{PCE}(\hat{\bt K}',T^{-1}_{\boldsymbol \alpha^*}(\bt W))$ can be expanded as follows
\begin{equation}
\text{ssq}(\langle \hat{\bt K}', T^{-1}_{\boldsymbol \alpha^*} (\tilde{\bt W}) \rangle) 
= \sum_{k=1}^L \text{ssq}(\mathsf{\Phi}_{k,\mathsf{inv}}(\alpha^*_k)),
\label{eq:es_num}
\end{equation}
Now we can easily identify each of the $L$ summands in \eqref{eq:es_num} as the numerator of $\varphi_{k,\mathsf{inv}}(\alpha^*_k)$ in \eqref{eq:varphi_inv} which can be stored during the adaptive optimization process. 

The denominator of the PCE requires more attention. With the approximation above, this denominator is $\frac{1}{|{\mathcal I_T} \backslash {\mathcal S}|}\sum_{\bt s \in {\mathcal I_T} \backslash {\mathcal S}} \langle  \hat{\bt K}',C(T^{-1}_{\boldsymbol \alpha^*} (\tilde{\bt W}), \bt s) \rangle^2$ which is nothing but a sample estimate of the variance of the cross-correlation of $\hat{\bt K}'$ and $T^{-1}_{\boldsymbol \alpha^*} (\tilde{\bt W})$. In \cite[Sect. VIII]{tech-report:Montibeller-Perez} we derive and discuss a simpler sample estimate that is more statistically efficient (i.e., has a lower variance). This fully justifies the approximation
\begin{equation}
\frac{1}{|{\mathcal I_T} \backslash {\mathcal S}|}\sum_{\bt s \in {\mathcal I_T} \backslash {\mathcal S}} \langle  \hat{\bt K}',C(T^{-1}_{\boldsymbol \alpha^*} (\tilde{\bt W}), \bt s) \rangle^2 \approx \kappa \cdot \hat \sigma_{\hat K'}^2 \cdot \sum_{k=1}^L \mathsf{E}_{k,\mathsf{inv}}(\alpha^*_k),
\label{eq:es_den}
\end{equation}
where $\hat \sigma_{\hat K'}^2 \doteq ||\hat{\bt K}'||^2/|{\mathcal I}|$ (recall that $\hat K'_{i,j}$ exists for all $(i,j) \in {\mathcal I}$), and $\kappa$ is a factor that takes into account the fact that the cardinalities of ${\mathcal I}$ and $\bigcup_{k=1}^L {\mathcal Q}_{k,\mathsf{inv}}(\alpha^*_k)$ are different. (In practice, $\kappa$ will be close to 1, so it can be dropped.) Once again, the $L$ summands in \eqref{eq:es_den} are already available as the denominator of \eqref{eq:varphi_inv}. 

\subsection{Early stopping}
\label{sec:early_stop}
The partition into annuli offers one remarkable byproduct:  taking inspiration from \cite{Perez2016Fast}, it is possible to stop processing annuli (and declare that $H_1$ holds) if a cumulative PCE exceeds a predefined threshold.  Following the approximations in the previous subsection, one might be tempted to compute a cumulative PCE by using the numerators and denominators already produced during the optimization. In this way,  the optimization would not need to be carried out for all annuli but instead it could be stopped as soon as the PCE computed so far exceeds the threshold. Unfortunately, this approach would be incorrect, because while a fraction with sums in the numerator can be expanded into a sum of fractions, this is not the case when there are sums in the denominator. Therefore, if we want to implement an early stopping mechanism, we need to seek ways to further approximate the denominator of the PCE without actually computing all the elements of $\boldsymbol \alpha^*$. To this end, we can ask ourselves how sensitive is the right hand side of \eqref{eq:es_den} to changes in $\alpha_k^*$; after all, since each of the $L$ summands is an estimate of the variance of the transformed residual inside an annulus, one would expect not much variation for realistic values of $\alpha$. If this were the case, then one might approximate the right hand side of \eqref{eq:es_den} (which corresponds to the optimal vector $\boldsymbol \alpha_\mathsf{inv}^*$) by computing it for any reasonable value of $\boldsymbol \alpha_\mathsf{inv}$ without involving any optimization. 

\begin{figure}[!ht]
\centering
\includegraphics[scale=0.65]{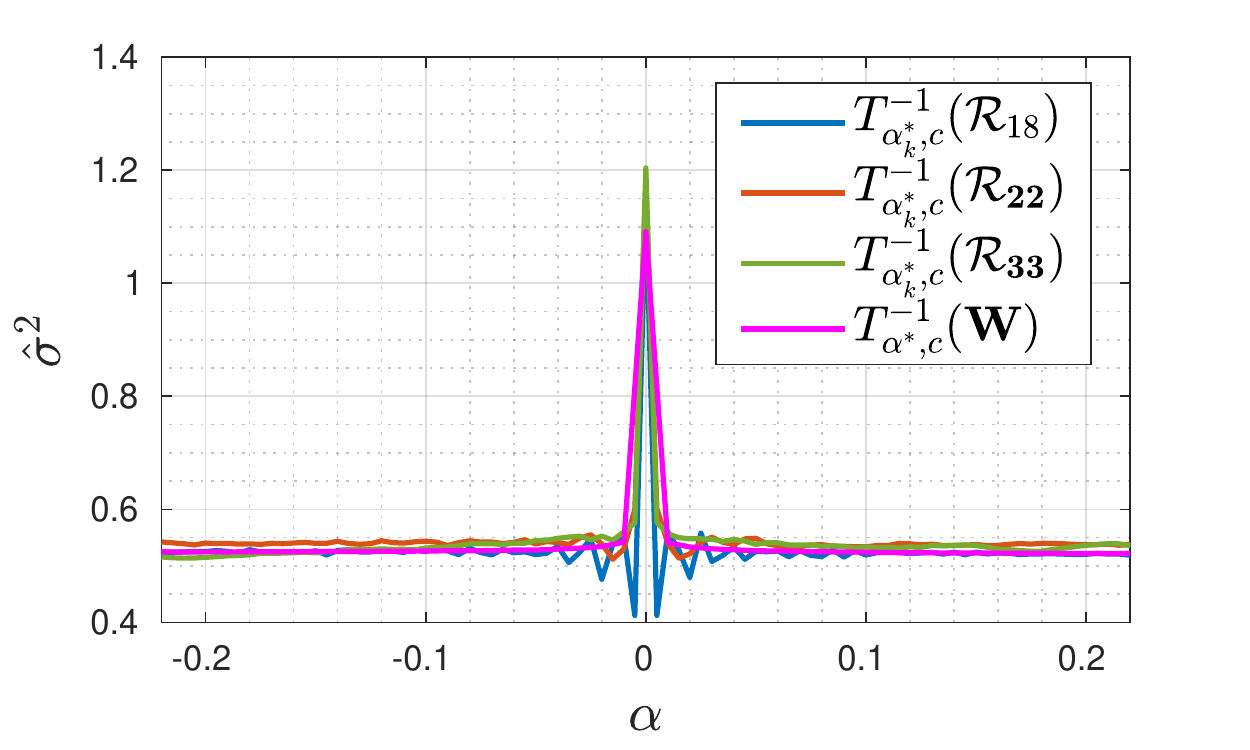}
\caption{Sample variance of the transformed
residual for different annuli and different values of $\alpha$. Camera and parameters are the same as in Fig.~\ref{fig:alpha_trend}.}
\label{fig:var_vs_alpha}
\end{figure}

In order to illustrate the feasibility of this approximation, we show in Fig.~\ref{fig:var_vs_alpha} the values of the sample variance of a transformed residual computed in each annulus, i.e., 
$\frac{\mathsf{E}_{k,\mathsf{inv}}(\alpha_k)}{|{\mathcal Q}_{k,\mathsf{inv}}(\alpha_k)|}$ as a function of $\alpha_k$ for several annuli (i.e., $k=18, 22, 33$) and for cubic inverse mappings, see \eqref{eq:inverse2}. Fig.~\ref{fig:var_vs_alpha} also shows the value of the variance estimated from the full-size transformed residual, i.e.,  $\frac{1}{|{\mathcal I}_T|} ||T^{-1}_{\alpha_k} (\tilde{\bt W})||^2$. Bi-linear interpolation is used in all cases.

As we can see, the variance estimate is fairly constant for different values of $\alpha_k$, except in a neighborhood of zero. Moreover, this is similar to the variance estimate obtained from the whole transformed residual, so the latter can be used in place of the variance estimate for a specific annulus. The reason for the spike at $\alpha_k=0$ is that the interpolation that is needed for computing the inverse mapping when $\alpha_k \neq 0$ produces a reduction in the variance of the transformed residual. This reduction depends on the square magnitude of the interpolation filter at different sampling points. In general, the grids before and after the interpolation are not related through rational numbers, but for certain rings and values of $\alpha$, moiré patterns between the sampling grids may appear; this is why in Fig.~\ref{fig:var_vs_alpha} a ripple near zero is observed for the rings $k=18, 22$.  The energy reduction phenomenon has been reported in  \cite{Goljan2018Blind} in a different scenario 
but related to ours.

The invariance discussed in the previous paragraph suggests several ways of approximating the right hand side of \eqref{eq:es_den}; for instance, it is possible to pick any value of $\alpha$, say $\alpha_\mathsf{f}$, sufficiently far from $\alpha=0$ and for all the annuli use the same transformation $T_{\alpha_\mathsf{f}}^{-1}$ in place of $T_{\alpha^*_k}^{-1}$. We remark that the reason why the neighborhood of $\alpha=0$ should be excluded when selecting $\alpha_{\mathsf{f}}$ is the fact that inside such a neighborhood the denominator of the PCE is overestimated and, consequently, the PCE underestimated.   

Another way of approximating the right hand side of \eqref{eq:es_den} which offers a slightly better performance than the former is to use the values of $\alpha_k^*$ already available from the optimization to update the approximation. This comes at practically no cost because the corresponding term $\mathsf{E}_{k,\mathsf{inv}}(\alpha_k^*)$ needs to be computed anyway during the optimization. For those annuli whose $\alpha_k^*$ is not available yet, the corresponding term is substituted by its approximation computed at $\alpha_k=\alpha_{\mathsf{f}}$.  

We explain next how to compute the Cumulative PCE at the $n$th iteration which we will denote by $\text{CPCE}_{n,\mathsf{inv}}(\hat{\bt K}',\bt W)$. First, we need a mapping $\xi: \{1, \cdots, L\} \to \{1, \cdots, L\}$, from the natural order to the one induced by the proposed iterative procedure, i.e., $\xi(1) \mapsto k_0, \xi(2) \mapsto k_0+1, \cdots, \xi(L-k_0) \mapsto L, \xi(L-k_0+1) \mapsto k_0-1, \cdots, \xi(L) \mapsto 1$. Then, 
\begin{equation}
    \text{CPCE}_{n,\mathsf{inv}}(\hat{\bt K}',\bt W) \doteq \frac{ \sum_{k=\xi(1)}^{\xi(n)} \text{ssq} \left(\mathsf{\Phi}_{k,\mathsf{inv}}(\alpha^*_k)\right)}
{\hat \sigma_{\hat K'}^2 \left( \sum\limits_{k=\xi(1)}^{\xi(n)} \hspace*{-0.2cm} \mathsf{E}_{k,\mathsf{inv}}(\alpha^*_k) + \hspace*{-0.2cm} \sum\limits_{k=\xi(n+1)}^{\xi(L)} \hspace*{-0.35cm} \mathsf{E}_{k,\mathsf{inv}}(\alpha^*_\mathsf{f})\right)}.
\end{equation}

Thus, the early-stopping algorithm will declare a match and stop if for some $n =1, \cdots, L$, $\text{CPCE}_{n,\mathsf{inv}}(\hat{\bt K}',\bt W) > \tau_c$ is satisfied. The value of $\tau_c$ is set experimentally to achieve the desired False Positive Rate (FPR). 

Given the numerator and denominator of $\text{CPCE}_{n,\mathsf{inv}}(\hat{\bt K}',\bt W)$, and once 
$\alpha_{\xi(n+1)}^*$ is available, the numerator of $\text{CPCE}_{n+1,\mathsf{inv}}(\hat{\bt K}',\bt W)$ is updated by adding $\mathsf{\Phi}_{\xi(n+1),\mathsf{inv}}(\alpha^*_{\xi(n+1)})$, while the update of the denominator requires adding $\mathsf{E}_{\xi(n+1),\mathsf{inv}}(\alpha^*_{\xi(n+1)})$ and subtracting $\mathsf{E}_{\xi(n+1),\mathsf{inv}}(\alpha_\mathsf{f})$. 

A similar definition follows for the Cumulative PCE in the direct approach $\text{CPCE}_{n,\mathsf{dir}}(\hat{\bt K}',\bt W)$ and the corresponding early stopping criterion.

\subsection{Parameter inheritance}
\label{sec:par_inh}

As we have discussed, the test decision statistic takes the maximum of the PCEs computed through the direct and the inverse approaches. This implies that it is necessary to compute the optimal vector $\boldsymbol \alpha^*$ for both approaches, so the computational complexity is roughly doubled. This also holds if the early stopping criterion introduced above is applied. In such a case, the iterations for both the direct and the inverse approaches are made in parallel, so that for every $k$ both  $\text{CPCE}_{n,\mathsf{dir}}(\hat{\bt K}',\bt W)$ and $\text{CPCE}_{n,\mathsf{inv}}(\hat{\bt K}',\bt W)$ are checked against the threshold in order to stop as soon as possible. 

There is one sub-optimal way to alleviate the computational burden due to keeping the two approaches. We term it {\em parameter inheritance} and basically consists in using for the direct approach the same vector $\boldsymbol \alpha^*$ that was computed for the inverse approach. Of course, the latter is not necessarily optimal for the direct approach, but the rationale is that inside each annulus ${\mathcal R}_k$ the direct and inverse transformations nearly correspond to each other for the same value of $\alpha_k$. Perfect correspondence does not exist because the inverse transformation is only an approximation and due to the fact that the search algorithm is prone to errors due to noise and insufficient resolution. 

\subsection{Parameter default values}
\label{sec:default}
In this section we provide the default values for the parameters of our algorithm and discuss some decisions regarding the initialization. These default values were used in the experiments reported in Sect.~\ref{sec:experiments}.  Specifically, the radius of the inner disk $r_1$ is such that $r_1 \cdot D_2$ equals 250 pixels and the width of each annulus $\Delta_k$ is such that $\Delta_k \cdot D_2$ equals 64 pixels. Both values are chosen as a compromise between performance and computational cost. For the linear predictor we set $U = 6$, $\mu = 1$ and $A_{\mathsf{min}}=7$. 

The initial search set  $\mathcal{A}_{k_0}$ is given by $\mathcal{A}_{k_0}=\{-0.22, -0.21, \cdots, 0.21, 0.22\}$, which is the same range as used and justified in  \cite{Goljian2012Sensor} to cover a variety of barrel and pincushion distortions.  However, in our case we apply a coarser resolution for computational reasons and because the adaptive nature of our algorithm automatically adjusts to finer resolutions after a few iterations. We are aware that in \cite{Goljian2014Estimation} a wider range was preferred (even if just to invert pincushion distortions), so we carried out some experiments with images taken with the Canon 1200D camera and radially corrected with Adobe Lightroom using the lens distortion model of a different device (see Section \ref{sec:experiments}), since this combination produces some of the strongest and most variable radial corrections of our dataset. In these experiments, the search set was expanded to $\mathcal{A}_{k_0}=\{-0.50, -0,49, \cdots, 0.49, 0.50\}$. While it is true that this set allows in some cases to get closer to the proper $\alpha_{k_0}$, we found no significant differences in terms of performance with the previous initialization; as mentioned, this is due to our algorithm quickly finding the right range for $\alpha_k$ after few iterations. In contrast, the computational load of using the enlarged search set would be larger; for this reason, we recommend $\mathcal{A}_{k_0}=\{-0.22, -0.21, \cdots, 0.21, 0.22\}$. For an in-depth complementary discussion on the initial set, please see \cite{tech-report:Montibeller-Perez}. 

After the initial search, for the forward prediction ${\mathcal A}_{k_0+1}$ is given by \eqref{eq:A_k} with $\lambda_{k_0+1} = 0.001$ and $A_{k_0+1}=9$. For the following iterations, 
\begin{equation}
\label{eq:cases}
\lambda_k= \begin{cases} 0.1 & \text{if  } |\alpha_k-\alpha_{k-1}| > 0.1, \\
0.01 & \text{if  } 0.01 < |\alpha_k-\alpha_{k-1}| \leq 0.1, \\
0.001 & \text{if  }  |\alpha_k-\alpha_{k-1}| \leq 0.001.
\end{cases}
\end{equation}

Identical considerations to the previous paragraph are made in regard to the backward prediction, where $k_0+1$ is replaced now by $k_0-1$ and in \eqref{eq:cases} $k-1$ is replaced by $k+1$.

\section{Experimental Results}
\label{sec:experiments}
In order to measure the performance of the methods presented in Sect.~\ref{sec:our_method} and compare them with the state of the art in \cite{Goljian2012Sensor} and \cite{Goljian2014Estimation}, we built a test dataset composed of 3645 images, of which 2037 were taken with the following compact cameras and radially corrected ``in-camera" (i.e., by the camera software): Canon SX230 HS (188 images), Panasonic ZS7 (170 images), Canon SX40 (57 images), Canon SX210 (82 images), and Nikon S9100 (1540 images). All these images were downloaded from Flickr, as done in \cite{Goljian2012Sensor} and \cite{Goljian2014Estimation}; for this reason, there is an uneven distribution of images per device. 1508 of the remaining images in the test dataset were taken with the Canon 1200D (a reflex camera not applying any type of in-camera post-processing) with the following Canon Zoom Lenses: 1) EF-S 10-18 mm 1:4-5.6 IS STM; 2) EF-S 18-55 mm 1:3.5-5.6; 3) EF 75-300 mm 1:4-5.6, all radially corrected ``out-camera" with third-party editing software: Adobe Lightroom Classic CC 2017, Adobe Photoshop CC 2017, PT Lens v2.0 (Macbook) and Gimp 2.10.14. Specifically, 377 images were corrected with each of these tools. With Adobe Lightroom we applied the correction model specific to the lens used to take the picture, thanks to the database of radial correction models Lightroom is equipped with. For the other editing software we applied the strongest radial correction available, as those tools cannot be tuned to a specific lens model.  The last 100 images in the test dataset were also taken with the Canon 1200D camera but corrected with Lightroom using models for other lenses (i.e., Nikon, Tamron, Apple, Huawei and DJI, with 20 images each), always applying the strongest radial correction. This latter subset will be labeled as ``Lightroom*'' in the following.  

Images in the test dataset were JPEG compressed with a QFs in the range 90-98. For each device, the same QF is consistently used; see \cite{tech-report:Montibeller-Perez} for details. The reference PRNUs for carrying out the tests were estimated for each device using \eqref{eq:camera_fingerprint} with $L=20$ natural images (not used for testing) compressed with matching QFs to the test subset of that device. For the compact devices, since the in-camera corrections depend on the focal length, fixed specific values of the latter were sought in order to estimate the respective PRNUs; whenever enough images were available for a certain device and focal length, a different fingerprint was estimated and the results averaged for each device. In all cases, hypothesis $H_1$ was tested with images taken with focal lengths different from those used to estimate the fingerprints. We refer the reader to \cite{tech-report:Montibeller-Perez}  for full details.  When, under hypothesis $H_0$, the test images and the fingerprints have different sizes, we crop the central part of the largest to match its size to the smallest \cite{Darvish2019Camera}. 

Next, we describe the identifiers used to refer to the different variants of our method in the figures and tables in this section. With ``Dir'' and ``Inv'' we indicate those cases where $\text{CPCE}_{n,\text{dir}}(\hat{\mathbf{K}^{\prime}}, \mathbf{W})$ and $\text{CPCE}_{n,\text{inv}}(\hat{\mathbf{K}^{\prime}}, \mathbf{W})$ are respectively used as the only test statistics. By ``2W'' we refer to the ``two-way'' case in which both the direct and the inverse approaches are used and $H_1$ is decided if either $\text{CPCE}_{n,\text{dir}}(\hat{\mathbf{K}^{\prime}}, \mathbf{W})$ or $\text{CPCE}_{n,\text{inv}}(\hat{\mathbf{K}^{\prime}}, \mathbf{W})$ are above the threshold for any $n \in \{1, \cdots, L\}$. To alleviate the computational load of the ``two-way'' parameter optimization, recall that in Sect.\ref{sec:par_inh} we proposed to inherit the parameters of one approach to the other. We will use the label $\overrightarrow{\text{DI}}$ to indicate inheritance of $\alpha_n^*$ from the direct approach to the inverse one; and $\overrightarrow{\text {ID}}$ vice versa. On the other hand, with the labels ``Cub" and ``Lin" we refer to the cubic and the linear radial correction models, respectively; see \eqref{eq:inverse2}. In all reported cases, the early stopping strategy from Sect.~\ref{sec:early_stop} is imposed. 

All the tests were run on a server with the following characteristics: 16 Cores, Processors 2xXeon E5-2667v3 3.2 GHz and RAM 192 GB; our implementation requires at most 5GB of RAM. In experimentally comparing the variants of our method with the algorithms proposed in \cite{Goljian2012Sensor} and \cite{Goljian2014Estimation}, we noticed that \cite{Goljian2012Sensor} was tested on images of size $3000 \times 4000$ that are, on average, larger than the in-camera corrected images in our dataset (refer to Table \ref{tab:time_consumed_and_tpr} for the image sizes in each subset). This explains the slightly worse performance measured here (with downsampling) compared to that reported in \cite{Goljian2012Sensor}.

In Table \ref{tab:table_roc} we provide the fixed thresholds $\tau_c$  (measured over the entire test dataset) that ensure False Positive Rates (FPR) of $0.05$ and $0.01$ together with the corresponding True Positive Rates (TPR) for the different variants of our method and those in \cite{Goljian2014Estimation} and  \cite{Goljian2012Sensor} (with and without DS). 

\begin{table}[!htbp]
\centering
\begin{tabular}{c|c|c|c|c}
                                               & $\tau_{0.05}$ & $\tau_{0.01}$ & TPR$_{0.05}$ & TPR$_{0.01}$ \\ \hline
$\overrightarrow{\text{ID}}$, Lin                             & 98.86         & 112.71        & 0.96         & 0.94         \\
$\overrightarrow{\text{DI}}$, Lin                                   & 90.01         & 105.63        & 0.97         & 0.95         \\
$\overrightarrow{\text{ID}}$, Cub                                    & 73.48         & 90.28         & 0.99         & 0.98         \\
$\overrightarrow{\text{DI}}$, Cub                                    & 71.13         & 84.75         & 0.99         & 0.98         \\
Inv, Lin                                       & 97.66         & 111.72        & 0.93         & 0.91         \\
Dir, Lin                                       & 90.01         & 105.63        & 0.95         & 0.92         \\
Inv, Cub                                       & 73.48         & 90.29         & 0.98         & 0.98         \\
Dir, Cub                                       & 71.13         & 84.57         & 0.99         & 0.98         \\
2W, Cub
& 71.12         & 84.34         & 0.99  & 0.99 \\
\cite{Goljian2012Sensor}                       & 4.81          & 8.36          & 0.86         & 0.82  \\
\cite{Goljian2012Sensor} no DS & 7.65          & 10.54         & 0.94         & 0.93         \\
\cite{Goljian2014Estimation}                       & 2.83          & 5.50          & 0.78         & 0.73            
\end{tabular}
\caption{Thresholds required to achieve FPR=0.05 and FPR=0.01 and corresponding TPRs for different methods and variants.} 
\label{tab:table_roc}
\end{table}

\begin{figure}[!htbp]
        \centering
		\includegraphics[scale=0.37]{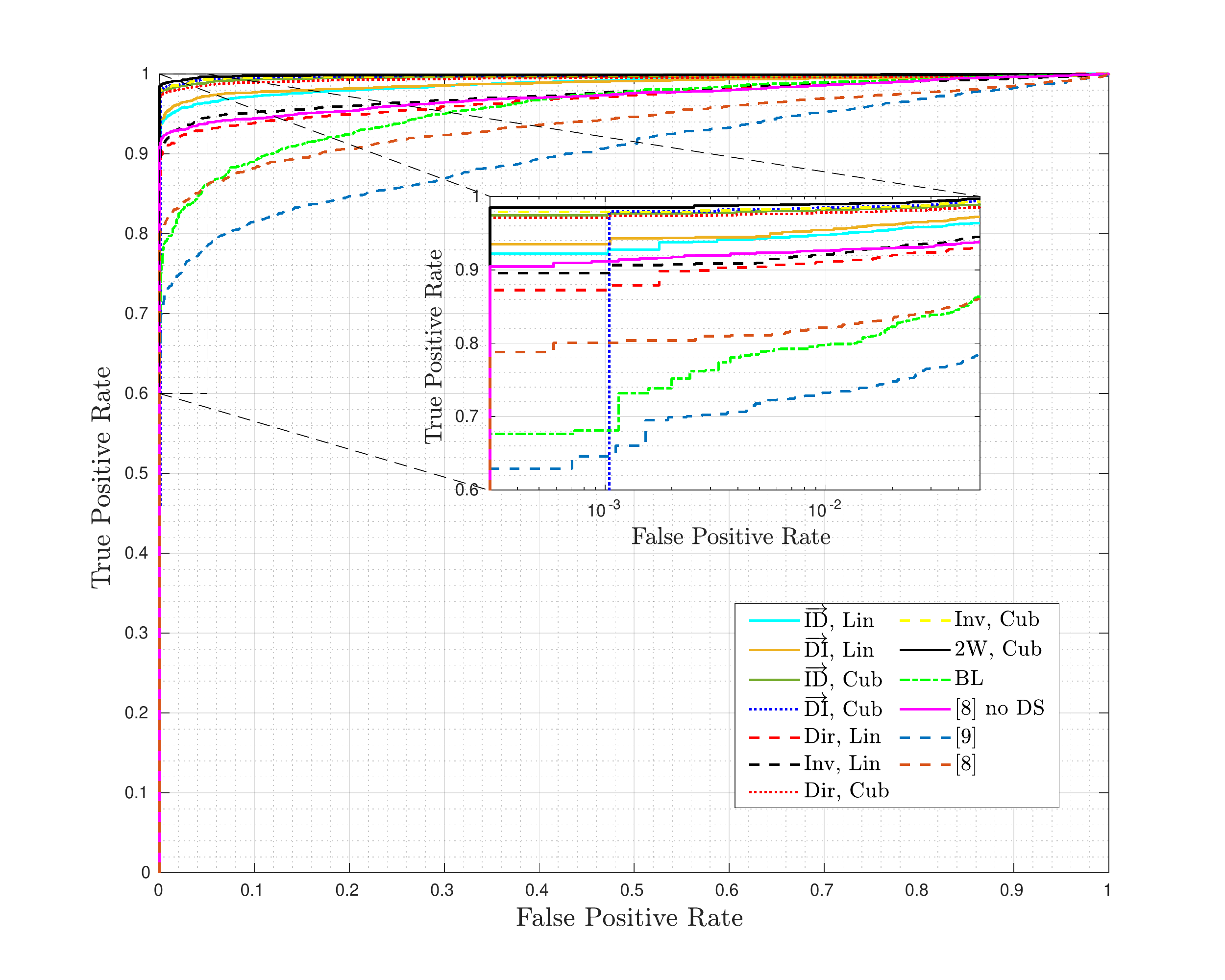}
		\caption{ROCs obtained with the variants of our method, \cite{Goljian2012Sensor} (with and without DS) and \cite{Goljian2014Estimation} on the test dataset. For a better visualization, the zoomed-in box corresponding to low FPRs uses a log scale on the x-axis.}  
		\label{fig:rocs}
\end{figure}

A breakdown by subset of the previous table is given in Table~\ref{tab:time_consumed_and_tpr} which also shows the time consumed to declare a match (under $H_1$) by the different alternatives (with early stopping in those cases where it applies).\footnote{The time consumed when no early stopping is in force is given in the technical report \cite{tech-report:Montibeller-Perez}.} For reasons of space, we have excluded the method in \cite{Goljian2014Estimation} which yields a modest performance, as well as the worst-performing variants of our method (cf. Table~\ref{tab:table_roc}); see \cite{tech-report:Montibeller-Perez} for fully comprehensive results. The Receiver Operating Characteristic (ROC) curves for the variants in Table~\ref{tab:time_consumed_and_tpr} are plotted in Fig.~\ref{fig:rocs}, where we have also added for comparison the baseline (BL) obtained by using $\text{PCE}(\hat{\mathbf{K}^{\prime}}, \mathbf{W})$ (i.e., with no transformations of either the PRNU or the residual) as test statistic. 

\begin{table*}[htp]
\resizebox{18cm}{!}{%
\begin{tabular}{c|cc|cc|cc|cc|cc|cc|cc|cc|cc|}
\multirow{2}{*}{}                                                                                  & \multicolumn{2}{c|}{$\overrightarrow{\text{ID}}$, Lin}     & \multicolumn{2}{c|}{$\overrightarrow{\text{DI}}$, Lin}     & \multicolumn{2}{c|}{$\overrightarrow{\text{ID}}$, Cub}             & \multicolumn{2}{c|}{$\overrightarrow{\text{DI}}$, Cub}              & \multicolumn{2}{c|}{Inv, Cub}                       & \multicolumn{2}{c|}{Dir, Cub}              & \multicolumn{2}{c|}{2W, Cub}  & \multicolumn{2}{c|}{\cite{Goljian2012Sensor} no DS} & \multicolumn{2}{c|}{\cite{Goljian2012Sensor}}     \\ \cline{2-19} 
                                                                                                   & \multicolumn{1}{c|}{TPR}  & time           & \multicolumn{1}{c|}{TPR}           & time  & \multicolumn{1}{c|}{TPR}           & time          & \multicolumn{1}{c|}{TPR}           & time           & \multicolumn{1}{c|}{TPR}           & time           & \multicolumn{1}{c|}{TPR}           & time  & \multicolumn{1}{c|}{TPR}           & time  & \multicolumn{1}{c|}{TPR}   & time   & \multicolumn{1}{c|}{TPR}  & time  \\ \hline
\multicolumn{1}{|c|}{\begin{tabular}[c]{@{}c@{}}GIMP\\ {[}$3456\times 5184${]}\end{tabular}}       & \multicolumn{1}{c|}{0.98} & 220.1          & \multicolumn{1}{c|}{0.98}          & 303.6 & \multicolumn{1}{c|}{0.98}          & 193.9         & \multicolumn{1}{c|}{\textbf{0.99}} & 204.8          & \multicolumn{1}{c|}{0.98}          & \textbf{182}   & \multicolumn{1}{c|}{0.98}          & 196.2 & \multicolumn{1}{c|}{\textbf{0.99}} & 326.4 & \multicolumn{1}{c|}{0.97}  & 962.3  & \multicolumn{1}{c|}{0.97} & 85.3  \\ \hline
\multicolumn{1}{|c|}{\begin{tabular}[c]{@{}c@{}}LIGHTROOM\\ {[}$3456\times 5184${]}\end{tabular}}  & \multicolumn{1}{c|}{0.93} & \textbf{237.8} & \multicolumn{1}{c|}{0.95}          & 330.3 & \multicolumn{1}{c|}{0.94}          & 288.4         & \multicolumn{1}{c|}{0.97}          & 309.1          & \multicolumn{1}{c|}{0.94}          & 274.6          & \multicolumn{1}{c|}{0.97}          & 295.1 & \multicolumn{1}{c|}{\textbf{0.98}} & 467.4 & \multicolumn{1}{c|}{0.6}   & 930.8  & \multicolumn{1}{c|}{0.44} & 87.9  \\ \hline
\multicolumn{1}{|c|}{\begin{tabular}[c]{@{}c@{}}LIGHTROOM*\\ {[}$3456\times 5184${]}\end{tabular}} & \multicolumn{1}{c|}{0.95} & 137.5          & \multicolumn{1}{c|}{0.92}          & 222.6 & \multicolumn{1}{c|}{\textbf{0.99}} & 158.8         & \multicolumn{1}{c|}{0.98}          & \textbf{126.5} & \multicolumn{1}{c|}{\textbf{0.99}} & 127.7          & \multicolumn{1}{c|}{0.98}          & 129.7 & \multicolumn{1}{c|}{0.98}          & 220.5 & \multicolumn{1}{c|}{0.51}  & 930.9  & \multicolumn{1}{c|}{0.4}  & 91.6  \\ \hline
\multicolumn{1}{|c|}{\begin{tabular}[c]{@{}c@{}}PHOTOSHOP\\ {[}$3456\times 5184${]}\end{tabular}}  & \multicolumn{1}{c|}{0.97} & \textbf{151.2} & \multicolumn{1}{c|}{\textbf{0.98}} & 159.5 & \multicolumn{1}{c|}{0.97}          & 165.4         & \multicolumn{1}{c|}{0.97}          & 176.2          & \multicolumn{1}{c|}{0.97}          & 170.7          & \multicolumn{1}{c|}{0.97}          & 167.3 & \multicolumn{1}{c|}{0.97}          & 293   & \multicolumn{1}{c|}{0.96}  & 974.6  & \multicolumn{1}{c|}{0.91} & 107.2 \\ \hline
\multicolumn{1}{|c|}{\begin{tabular}[c]{@{}c@{}}PT LENS\\ {[}$3456\times 5184${]}\end{tabular}}    & \multicolumn{1}{c|}{0.97} & \textbf{181.3} & \multicolumn{1}{c|}{0.99}          & 247.3 & \multicolumn{1}{c|}{0.98}          & 248.9         & \multicolumn{1}{c|}{0.99}          & 253.6          & \multicolumn{1}{c|}{0.98}          & 256.3          & \multicolumn{1}{c|}{\textbf{1}}    & 238.2 & \multicolumn{1}{c|}{\textbf{1}}    & 412.4 & \multicolumn{1}{c|}{0.91}  & 918.3  & \multicolumn{1}{c|}{0.8}  & 100.3 \\ \hline
\multicolumn{1}{|c|}{\begin{tabular}[c]{@{}c@{}}S9100\\ {[}$3000\times4000${]}\end{tabular}}       & \multicolumn{1}{c|}{0.99} & 114.2          & \multicolumn{1}{c|}{0.98}          & 117.3 & \multicolumn{1}{c|}{0.99}          & \textbf{72.9} & \multicolumn{1}{c|}{0.99}          & 99.8           & \multicolumn{1}{c|}{0.99}          & 92.9           & \multicolumn{1}{c|}{\textbf{1}}    & 92.5  & \multicolumn{1}{c|}{\textbf{1}}    & 160.3 & \multicolumn{1}{c|}{1}     & 598.3  & \multicolumn{1}{c|}{0.98} & 81.7  \\ \hline
\multicolumn{1}{|c|}{\begin{tabular}[c]{@{}c@{}}SX210\\ {[}$3240\times4320${]}\end{tabular}}       & \multicolumn{1}{c|}{0.96} & 256.8          & \multicolumn{1}{c|}{0.97}          & 262.5 & \multicolumn{1}{c|}{1}             & 127.5         & \multicolumn{1}{c|}{1}             & 139.1          & \multicolumn{1}{c|}{\textbf{1}}    & \textbf{121.8} & \multicolumn{1}{c|}{\textbf{1}}    & 126.8 & \multicolumn{1}{c|}{\textbf{1}}    & 210.3 & \multicolumn{1}{c|}{1}     & 723.4  & \multicolumn{1}{c|}{0.99} & 81.3  \\ \hline
\multicolumn{1}{|c|}{\begin{tabular}[c]{@{}c@{}}SX230\\ {[}$1584\times2816${]}\end{tabular}}       & \multicolumn{1}{c|}{0.87} & 33.8           & \multicolumn{1}{c|}{0.9}           & 29    & \multicolumn{1}{c|}{1}             & \textbf{23}   & \multicolumn{1}{c|}{1}             & 24.2           & \multicolumn{1}{c|}{\textbf{0.98}} & 41             & \multicolumn{1}{c|}{\textbf{0.98}} & 41.8  & \multicolumn{1}{c|}{\textbf{0.98}} & 70.5  & \multicolumn{1}{c|}{0.98}  & 229.0  & \multicolumn{1}{c|}{0.83} & 25.1  \\ \hline
\multicolumn{1}{|c|}{\begin{tabular}[c]{@{}c@{}}SX40\\ {[}$2664\times4000${]}\end{tabular}}        & \multicolumn{1}{c|}{1}    & 105.1          & \multicolumn{1}{c|}{0.96}          & 145   & \multicolumn{1}{c|}{1}             & 93.4          & \multicolumn{1}{c|}{1}             & 103.4          & \multicolumn{1}{c|}{\textbf{1}}    & \textbf{88.6}  & \multicolumn{1}{c|}{\textbf{1}}    & 93.6  & \multicolumn{1}{c|}{\textbf{1}}    & 154   & \multicolumn{1}{c|}{1}     & 526.3  & \multicolumn{1}{c|}{0.98} & 56.3  \\ \hline
\multicolumn{1}{|c|}{\begin{tabular}[c]{@{}c@{}}ZS7\\ {[}$1920\times2560${]}\end{tabular}}         & \multicolumn{1}{c|}{0.76} & 42.7           & \multicolumn{1}{c|}{0.81}          & 35    & \multicolumn{1}{c|}{0.98}          & \textbf{27.7} & \multicolumn{1}{c|}{0.98}          & 29.8           & \multicolumn{1}{c|}{0.93}          & 50.5           & \multicolumn{1}{c|}{\textbf{0.99}} & 46.9  & \multicolumn{1}{c|}{\textbf{0.99}} & 79.6  & \multicolumn{1}{c|}{0.98}  & 255.2  & \multicolumn{1}{c|}{0.79} & 27.8  \\ \hline
\end{tabular}}
\caption{TPRs of the different variants of our method and \cite{Goljian2012Sensor} (with and without DS), and
the average time consumed to declare a match for specific devices/software. In \textbf{bold} we highlight the fastest and most accurate variants of our method.}
\label{tab:time_consumed_and_tpr}
\end{table*}

From the results in Fig.~\ref{fig:rocs} and Table \ref{tab:table_roc}, it is possible to conclude that the best performing variants of our method correspond to the cubic correction model (``Cub''), with ``$\overrightarrow{\text{DI}}$'', ``Dir'', ``Inv'' and ``2W'' all achieving similar TPRs for the target FPRs. On the other hand, the average execution time of the ``one way'' variants, i.e. ``Dir'' and ``Inv'', is lower because only one statistic has to be computed per iteration. The experiments, conducted on a dataset composed of a variety of radial corrections, show that our variants outperform \cite{Goljian2012Sensor} (both with or without DS) in terms of TPR. Moreover, thanks to our early stopping strategy, our fastest versions (i.e., ``Dir, Cub'' and ``Inv, Cub'') achieve under $H_1$ execution times that are comparable to \cite{Goljian2012Sensor} with DS. We also note that the original solution proposed in \cite{Goljian2012Sensor} (i.e. with DS) achieves a limited performance both on low-resolution devices (i.e. SX230 and ZS7) and in presence of complex out-camera radial corrections as those applied by Adobe Lightroom. Adapting \cite{Goljian2012Sensor} to avoid DS results in a significant performance increase in those difficult cases, at the expense of a much more costly execution. Nevertheless, for some severe radial corrections like those in our ``Lightroom*'' subset, using the full resolution in \cite{Goljian2012Sensor} is still not sufficient. In contrast, our method is able to adapt to this high complexity and offers an excellent performance with an affordable execution time.  

\section{Conclusions}
\label{sec:conclusions}
In this paper, we have proposed an adaptive method for PRNU-based camera attribution that is able to cope with complex radial distortion corrections, as those performed in-camera by most compact models and out-camera by image processing software. Existing approaches try to either ``correct'' the reference fingerprint or invert the correction by applying a further geometric transformation that, in order to avoid a combinatorial explosion, must use a reduced number of parameters. In turn, this limitation accounts for unsatisfactory performance when complex radial distortion corrections are in effect, an undesirable aspect in view of the trend of more elaborate transformations that are made possible by ever more powerful distortion correction firmware/software. Our approach is radically different: by applying a divide-and-conquer principle, embodied in the use of annuli, we are able to: 1) allow for complex distortion corrections, as locally the transformation undergone by each annulus is much simpler; 2) implement an early stopping strategy that offers large computational savings. The results presented in the paper clearly reveal that our algorithm (in most of its variants) outperforms the state of the art when accuracy and computational load are considered.   

We believe that the adaptive approach proposed here could also be fruitful in other very challenging camera attribution scenarios with a number of latent parameters, such as in HDR images \cite{Darvish2019Camera}, in-camera-stabilized videos \cite{mandelli2019facing}, and emerging in-camera processing \cite{iuliani2021leak}.

\begin{appendices}

\section{Derivation of the estimator of $\alpha_k^*$}
\label{app:estimator_alpha}
In this Appendix we derive a plausible estimator of $\alpha_k^*$ under the inverse approach; the derivation would be identical for the direct approach and, hence, is skipped here.  See the definition of  $\alpha_k^*$ in \eqref{eq:argmax_i}. We introduce a super-index in $\alpha_k$ to enumerate the elements of the candidate set ${\mathcal A}_k$, i.e., $\{\alpha_k^{(n)}: n=1, \cdots, A_k\}={\mathcal A_k}$.  

We assume that $H_1$ holds, i.e., $\bt I$ contains $\bt K'$, and the following model for the residuals:
\begin{equation}
\label{eq:res_model}
[T^{-1}_{\alpha_k^{(n)}}(\tilde{\bt W})]_{i,j}=\gamma_{i,j}^{(n)} [T^{-1}_{\alpha^{(n)}_k}(T_{\alpha^\dagger_k}(\hat{\bt K}'))]_{i,j} + N_{i,j}^{(n)}
\end{equation}
for all $(i,j) \in {\mathcal Q}_{k,\mathsf{inv}}(\alpha_k^{(n)})$ and $n \in \{1, \cdots, A_k\}$. In \eqref{eq:res_model} $\alpha^\dagger_k$ represents the {\em true} (locally for the $k$th annulus) value of $\alpha$. The multipliers $\gamma_{i,j}^{(n)}$ are non-negative and take into account both the multiplicative effect of the image $\bt I$ and the gain of the effective denoising filter (which also impacts on the estimate $\hat{\bt K}'$ of the true PRNU). We argue that these multipliers are very hard to estimate accurately; as a consequence, a full maximum likelihood decision will not be possible and some simplifications will be required. One such simplification is to consider that the cross-correlations between $\hat{\bt K}'$ and $T^{-1}_{\alpha_k^{(n)}}(\tilde{\bt W})$, for all  $n =1, \cdots, A_k$, constitute a set of sufficient statistics for the estimation problem. Recall from \eqref{eq:Phi} that these cross-correlations are denoted by $\mathsf{\Phi}_{k,\mathsf{inv}}(\alpha^{(n)}_k)$.

We make the following hypotheses:\\
\noindent
1) {\em Spikiness:}  The $\alpha_k^{(n)}$ are sufficiently separated so that the $\mathsf{\Phi}_{k,\mathsf{inv}}(\alpha^{(n)}_k)$ are mutually uncorrelated and $\mathbb{E}\{\mathsf{\Phi}_{k,\mathsf{inv}}(\alpha^{(n)}_k)\}=0$, for all $n =1, \cdots, A_k$, except for $n = l$, where $l$ is such that $\alpha^{(l)}_k$ is the closest to the true value $\alpha^\dagger_k$ and the expectation is taken over the underlying distribution of $\bt K'$. This hypothesis is reasonable in view of the spikiness of the PCE with $\alpha$ (see Fig.~\ref{fig:spiky_PCE}). We also assume that $\alpha^{(l)}_k$ is close enough to $\alpha^\dagger_k$ so that $[T^{-1}_{\alpha^{(l)}_k}(T_{\alpha^\dagger_k}(\hat{\bt K}'))]_{i,j} \approx \hat{K}'_{i,j}$ for all $(i,j) \in {\mathcal Q}_{k,\mathsf{inv}}(\alpha_k^{(l)})$.\\
\noindent
2) {\em Uncorrelatedness:} In \eqref{eq:res_model}, $N_{i,j}^{(n)}$ and $\gamma_{i,j}^{(n)} [T^{-1}_{\alpha^{(l)}_k}(\hat{\bt K}')]_{i,j}$ are zero-mean and mutually uncorrelated for all $(i,j) \in {\mathcal Q}_{k,\mathsf{inv}}(\alpha_k^{(n)})$ and $n \in \{1, \cdots, A_k\}$. For any $l, n \in \{1, \cdots, A_k\}$, $l \neq n$, the variables $\hat K'_{i,j} \cdot N_{i,j}^{(n)}$ and $\hat K'_{u,v} \cdot N_{u,v}^{(l)}$ are mutually uncorrelated for every $(i,j) \in {\mathcal Q}_{k,\mathsf{inv}}(\alpha_k^{(n)})$ and every $(u,v) \in {\mathcal Q}_{k,\mathsf{inv}}(\alpha_k^{(l)})$. \\
3) {\em Weak PRNU:}  In \eqref{eq:res_model}, $\left|\gamma_{i,j}^{(n)}  [T^{-1}_{\alpha^{(n)}_k}(T_{\alpha^\dagger_k}(\hat{\bt K}'))]_{i,j} \right| \ll |N_{i,j}|$ for a large number of pixels of each annulus; we write this more precisely as 
\begin{equation}
\sum_{(i,j) \in {\mathcal Q}_{k,\mathsf{inv}}(\alpha_k^{(n)})}  \left(\gamma_{i,j}^{(n)}\right)^2 [T^{-1}_{\alpha^{(n)}_k}(T_{\alpha^\dagger_k}(\hat{\bt K}'))]_{i,j}^2 \ll \mkern-36mu
 \sum_{(i,j) \in {\mathcal Q}_{k,\mathsf{inv}}(\alpha_k^{(n)})} \mkern-36mu  N_{i,j}^2
\end{equation}
for all $n \in \{1, \cdots, A_k\}$.

As a consequence of the spikiness and uncorrelatedness assumptions above and the Central Limit Theorem (which is applicable if we assume that $|{\mathcal Q}_{k,\mathsf{inv}}(\alpha_k^{(n)})|$ is large for all  $n \in \{1, \cdots, A_k\}$), the variables $\mathsf{\Phi}_{k,\mathsf{inv}}(\alpha^{(n)}_k)$ will be well modeled by independent Gaussian distributions, so the cross-correlations will be 
\begin{eqnarray}
\label{eq:the_phis}
\mathsf{\Phi}_{k,\mathsf{inv}}(\alpha^{(n)}_k) &\sim& {\mathcal N}(0, (\sigma^{(n)})^2),\   n=1, \cdots, A_k, n \neq l\\
\mathsf{\Phi}_{k,\mathsf{inv}}(\alpha^{(l)}_k) &\sim& {\mathcal N}(\mu^{(l)}, (\sigma^{(l)})^2) 
\end{eqnarray}
where ${\mathcal N}(\mu, \sigma^2)$ denotes a Gaussian with mean $\mu$ and variance $\sigma^2$, $\mu^{(l)}$ denotes the expected value of the cross-correlation for the value of $\alpha^{(n)}_k \in {\mathcal A}_k$ that is closest to $\alpha^\dagger_k$, and $(\sigma^{(n)})^2$, $n=1, \cdots, A_k$, denote the variances of the cross-correlations. 

For all $n=1, \cdots, A_k$, the variances $(\sigma^{(n)})^2$ can be written as $\text{Var} \{\sum_{(i,j) \in  {\mathcal Q}_{k,\mathsf{inv}}(\alpha_k)} \hat K'_{i,j} \cdot N_{i,j} \} \approx \hat \sigma_{\hat K'}^2 \sum_{(i,j) \in  {\mathcal Q}_{k,\mathsf{inv}}(\alpha_k)} N^2_{i,j}$. As a consequence of the weak PRNU assumption $\sum_{(i,j) \in  {\mathcal Q}_{k,\mathsf{inv}}(\alpha_k)} N^2_{i,j} \approx \mathsf{E}_{k,\mathsf{inv}}(\alpha_k^{(n)})$. Therefore, $(\sigma^{(n)})^2 \approx \hat \sigma_{\hat K'}^2 \mathsf{E}_{k,\mathsf{inv}}(\alpha_k^{(n)})$.  

Let $f_{\mathcal N}(Y;\mu,\sigma)$ denote the Gaussian pdf on random variable $Y \sim {\mathcal N}(\mu,\sigma^2)$. Also, let $\mathcal{E}_l$ denote the event  ``$\alpha^{(l)}_k$,  $l \in \{1, \cdots, A_k\}$ is the closest to the true value $\alpha^\dagger_k$''. Then the likelihood of jointly observing the cross-correlations $\mathsf{\Phi}_{k,\mathsf{inv}}(\alpha^{(n)}_k)$ conditioned on ${\mathcal E}_l$ is 
\begin{eqnarray}
\label{eq:likelihood}
f(\mathsf{\Phi}_{k,\mathsf{inv}}(\alpha^{(1)}_k), \cdots, \mathsf{\Phi}_{k,\mathsf{inv}}(\alpha^{(A_k)}_k)|{\mathcal E}_l) = \nonumber \\
f_{\mathcal N}(\mathsf{\Phi}_{k,\mathsf{inv}}(\alpha^{(l)}_k); \mu^{(l)}, \sigma^{(l)}) \cdot \prod_{\stackrel{n=1}{n\neq l}}^{A_k}  f_{\mathcal N}(\mathsf{\Phi}_{k,\mathsf{inv}}(\alpha^{(l)}_k); 0, \sigma^{(n)})
\end{eqnarray} 
The maximum likelihood estimator would be obtained by maximizing the likelihood in \eqref{eq:likelihood} with respect to $l$. The estimator will not change if we divide \eqref{eq:likelihood} by $\prod_{n=1}^{A_k} f_{\mathcal N}(\mathsf{\Phi}_{k,\mathsf{inv}}(\alpha^{(n)}_k); 0, \sigma^{(n)})$; this gives the following simpler likelihood function
\begin{eqnarray}
L(\mathsf{\Phi}_{k,\mathsf{inv}}(\alpha^{(1)}_k), \cdots, \mathsf{\Phi}_{k,\mathsf{inv}}(\alpha^{(A_k)}_k)|{\mathcal E}_l) \nonumber \\
=\frac{f_{\mathcal N}(\mathsf{\Phi}_{k,\mathsf{inv}}(\alpha^{(l)}_k); \mu^{(l)}, \sigma^{(l)})}{f_{\mathcal N}(\mathsf{\Phi}_{k,\mathsf{inv}}(\alpha^{(l)}_k); 0, \sigma^{(l)})}
\end{eqnarray}
Taking the logarithm and simplifying, we find that the maximum likelihood estimator is equivalent to solving 
\begin{equation}
\label{eq:first_estimator}
l^* = \arg \max_{l=1, \cdots, A_k} \mathsf{\psi}^{(l)}
\end{equation}
where
\begin{equation}
\label{eq:xi}
\mathsf{\psi}^{(l)} \doteq \frac{\mu^{(l)} \cdot \mathsf{\Phi}_{k,\mathsf{inv}}(\alpha^{(l)}_k)}{(\sigma^{(l)})^2} - \frac{1}{2} \frac{(\mu^{(l)})^2}{(\sigma^{(l)})^2}
\end{equation}
and making $\alpha^*_k = \alpha_k^{(l^*)}$. 

Notice that, as discussed above, $(\sigma^{(l)})^2$ can be replaced by its estimator $\hat \sigma_{\hat K'}^2 \mathsf{E}_{k,\mathsf{inv}}(\alpha_k^{(l)})$ in \eqref{eq:xi}. Unfortunately, producing a reliable estimator of $\mu^{(l)}$ is not feasible due to the unavailability of the gains $\gamma_{i,j}^{(l)}$. For this reason, we turn our attention to suboptimal estimators that can be practically implemented. If we assume that for all $l$ in $\{1, \cdots, A_k\}$ both $\mu^{(l)}$ and the ratio $\mu^{(l)}/\sigma^{(l)}$ do not vary significantly around their respective means, we can think of replacing $\mu^{(l)}$ and $\mu^{(l)}/\sigma^{(l)}$ in \eqref{eq:xi} by those means. This yields the simplified functional
\begin{equation}
\label{eq:xi2}
\mathsf{\psi}'^{(l)} \doteq {\mathsf{\Phi}_{k,\mathsf{inv}}(\alpha^{(l)}_k)}/{(\sigma^{(l)})^2}
\end{equation}
to be used in \eqref{eq:first_estimator}. After replacing $(\sigma^{(l)})^2$ in \eqref{eq:xi2} by its estimator $\hat \sigma_{\hat K'}^2 \mathsf{E}_{k,\mathsf{inv}}(\alpha_k^{(l)})$, and dropping $\hat \sigma_{\hat K'}^2$  because it is independent of $l$, we obtain the proposed \eqref{eq:varphi_inv}. 

It is interesting to evaluate the loss of performance that results when using \eqref{eq:xi2} instead of \eqref{eq:xi}. We do so by assuming w.l.o.g. that ${\mathcal E}_l$ holds and estimate the probabilities that a given $n \in \{1, \cdots, A_k\}$, $n \neq l$, produces a larger value than for $n=l$ in $\psi^{(n)}$ and $\psi'^{(n)}$. Then, we compare the two resulting probabilities in terms of the effective signal-to-noise ratios (SNR).  Therefore, in this case, following \eqref{eq:the_phis}, we have that for $n \neq l$, $\mathsf{\Phi}_{n,\mathsf{inv}}(\alpha^{(n)}_k) \sim {\mathcal N}(0,(\sigma^{(n)})^2)$, and $\mathsf{\Phi}_{k,\mathsf{inv}}(\alpha^{(l)}_k) \sim {\mathcal N}(\mu^{(l)},(\sigma^{(l)})^2)$. Thus, when ${\mathcal E}_l$ holds, $\psi^{(l)} \sim {\mathcal N}\left( (\mu^{(l)})^2/(\sqrt{2} \sigma^{(l)})^2, (\mu^{(l)})^2/(\sigma^{(l)})^2\right)$ and $\psi^{(n)} \sim {\mathcal N}\left(-(\mu^{(n)})^2/(\sqrt{2} \sigma^{(n)})^2,(\mu^{(n)})^2/(\sigma^{(n)})^2\right)$, $n \neq l$. Since $\psi^{(l)}$ and $\psi^{(n)}$ are independent, the probability that $\psi^{(n)} \geq \psi^{(l)}$ when ${\mathcal E}_l$ holds is the probability that the random variable $\psi^{(l)}-\psi^{(n)}$ is less than zero. And since both variables are Gaussian, so is their difference. Therefore, $
\psi^{(l)}-\psi^{(n)} \sim {\mathcal N}(\omega_{n,l}/2,\omega_{n,l})$, where 
\begin{equation}
\omega_{n,l} \doteq \frac{(\mu^{(l)})^2}{(\sigma^{(l)})^2} + \frac{(\mu^{(n)})^2}{(\sigma^{(n)})^2}
\end{equation}
If we define the effective SNR as the ratio between the squared mean and the variance of $\psi^{(l)}-\psi^{(n)}$, then we find that $\text{SNR}_{\psi}=\omega_{n,l}/4$, where the subindex $\psi$ indicates that we are using the estimator in \eqref{eq:xi}. 

For the simplified estimator in \eqref{eq:xi2}, a similar derivation leads to showing that
\begin{equation}
\psi'^{(l)}-\psi'^{(n)} \sim {\mathcal N}\left(\frac{\mu^{(l)}}{(\sigma^{(l)})^2}, \left[ \frac{1}{(\sigma^{(l)})^2} + \frac{1}{(\sigma^{(n)})^2}\right] \right)
\end{equation} 
for which the effective SNR, denoted as $\text{SNR}_{\psi'}$ is now
\begin{equation}
\text{SNR}_{\psi'} = \frac{(\mu^{(l)})^2/(\sigma^{(l)})^2}{\frac{(\sigma^{(l)})^2}{(\sigma^{(n)})^2} +1}
\end{equation}
In order to compare the effective SNRs, we compute their ratio:
\begin{equation}
\label{eq:ratio_SNR}
\frac{\text{SNR}_{\psi}}{\text{SNR}_{\psi'}}=\frac{1+\left(\frac{\mu^{(n)}}{\mu^{(l)}}\right)^2 \cdot \left(\frac{\sigma^{(l)}}{\sigma^{(n)}}\right)^2}{4} \cdot \left( \left(\frac{\sigma^{(l)}}{\sigma^{(n)}}\right)^2 +1 \right)
\end{equation}
To get a cleaner interpretation of this result, we can further assume that both $\mu^{(n)}$ and $(\sigma^{(n)})^2$ are proportional to the cardinality of the support set $|{\mathcal Q}_{k,\mathsf{inv}}(\alpha_k^{(n)})|$. This way, if we let $\beta_{n,l} \doteq  |{\mathcal Q}_{k,\mathsf{inv}}(\alpha_k^{(n)})|/ |{\mathcal Q}_{k,\mathsf{inv}}(\alpha_k^{(l)})|$, we can write that $\mu^{(n)}/\mu^{(l)}=\beta_{n,l}$ and  $(\sigma^{(l)})^2/(\sigma^{(n)})^2 = \beta_{n,l}^{-1}$. Then, substituting into \eqref{eq:ratio_SNR} we find that
\begin{equation}
\frac{\text{SNR}_{\psi}}{\text{SNR}_{\psi'}}=\frac{(1+\beta_{n,l})^2}{4 \beta_{n,l}}=1+\frac{(1-\beta_{n,l})^2}{4 \beta_{n,l}}
\end{equation}
which is clearly larger than one for all $\beta_{n,l} \geq 0$, $\beta_{n,l} \neq 1$. This confirms that, as expected, for any $\beta_{n,l} \neq 1$ there is a loss of effective SNR with respect to the optimal estimator. However, in practice this loss will be rather small: for instance, suppose that $|{\mathcal Q}_{k,\mathsf{inv}}(\alpha_k^{(n)})|$ is within the range of 20\% larger and 20\% smaller than $|{\mathcal Q}_{k,\mathsf{inv}}(\alpha_k^{(l)})|$, then the effective SNR for the suboptimal detector is at most 0.054 dB smaller than the corresponding to the optimal one. 

This supports the plausibility of the proposed simplified detector.

\end{appendices}
\vspace{-10pt}
\medskip
\bibliographystyle{ieeetr}
\bibliography{Biblio}

\begin{thebibliography}{10}

\bibitem{Lukas2006}
J.~Lukas, J.~Fridrich, and M.~Goljan, ``Digital camera identification from
  sensor pattern noise,'' {\em IEEE Transactions on Information Forensics and
  Security}, vol.~1, no.~2, pp.~205--214, 2006.

\bibitem{Rosenfeld:2009}
K.~Rosenfeld and H.~T. Sencar, ``A study of the robustness of prnu-based camera
  identification,'' in {\em Media Forensics and Security}, vol.~7254,
  p.~72540M, International Society for Optics and Photonics, 2009.

\bibitem{Taspinar2020}
S.~Taspinar, M.~Mohanty, and N.~Memon, ``Camera fingerprint extraction via
  spatial domain averaged frames,'' {\em IEEE Transactions on Information
  Forensics and Security}, vol.~15, pp.~3270--3282, 2020.

\bibitem{Korus:2017}
P.~{Korus} and J.~{Huang}, ``"{Multi-Scale Analysis Strategies in PRNU-Based
  Tampering Localization}",'' {\em IEEE Transactions on Information Forensics
  and Security}, vol.~12, no.~4, pp.~809--824, 2017.

\bibitem{Goljan2008Digital}
M.~Goljan, ``Digital camera identification from images--estimating false
  acceptance probability,'' in {\em International workshop on digital
  watermarking}, pp.~454--468, Springer, 2008.

\bibitem{Taspinar:2016}
S.~Taspinar, M.~Mohanty, and N.~Memon, ``Source camera attribution using
  stabilized video,'' in {\em 2016 IEEE International Workshop on Information
  Forensics and Security (WIFS)}, pp.~1--6, IEEE, 2016.

\bibitem{Darvish2019Camera}
D.~Morshedi, M.~Hosseini, and M.~Goljan, ``Camera identification from hdr
  images,'' in {\em Proceedings of the ACM Workshop on Information Hiding and
  Multimedia Security}, pp.~69--76, 2019.

\bibitem{Goljian2012Sensor}
M.~Goljan and J.~Fridrich, ``Sensor-fingerprint based identification of images
  corrected for lens distortion,'' in {\em Media Watermarking, Security, and
  Forensics 2012}, vol.~8303, p.~83030H, International Society for Optics and
  Photonics, 2012.

\bibitem{Goljian2014Estimation}
M.~Goljan and J.~Fridrich, ``Estimation of lens distortion correction from
  single images,'' in {\em Media Watermarking, Security, and Forensics 2014},
  vol.~9028, p.~90280N, International Society for Optics and Photonics, 2014.

\bibitem{Chen2008Determining}
M.~Chen, J.~Fridrich, M.~Goljan, and J.~Luk{\'a}s, ``Determining image origin
  and integrity using sensor noise,'' {\em IEEE Transactions on information
  forensics and security}, vol.~3, no.~1, pp.~74--90, 2008.

\bibitem{Mihcak1999Denoiser}
M.~K. Mihcak, I.~Kozintsev, K.~Ramchandran, and P.~Moulin, ``Low-complexity
  image denoising based on statistical modeling of wavelet coefficients,'' {\em
  IEEE Signal Processing Letters (SPL)}, vol.~6, pp.~300--303, 1999.

\bibitem{Goljan2009Large}
M.~Goljan, J.~Fridrich, and T.~Filler, ``Large scale test of sensor fingerprint
  camera identification,'' {\em Proceedings of SPIE - The International Society
  for Optical Engineering}, February 2009.

\bibitem{Kang12}
X.~Kang, Y.~Li, Z.~Qu, and J.~Huang, ``Enhancing source camera identification
  performance with a camera reference phase sensor pattern noise,'' {\em IEEE
  Transactions on Information Forensics and Security}, vol.~7, no.~2,
  pp.~393--402, 2012.

\bibitem{Hugemann2010Correcting}
W.~Hugemann, ``Correcting lens distortions in digital photographs,'' {\em
  Ingenieurb{\"u}ro Morawski+ Hugemann: Leverkusen, Germany}, vol.~20, 2010.

\bibitem{Li2005nonIterative}
H.~Li and R.~Hartley, ``A non-iterative method for correcting lens distortion
  from nine point correspondences,'' {\em OMNIVIS 2005}, vol.~2, p.~7, 2005.

\bibitem{Janez2002Nonparametric}
J.~Per{\v{s}} and S.~Kovacic, ``Nonparametric, model-based radial lens
  distortion correction using tilted camera assumption,'' in {\em Proceedings
  of the Computer Vision Winter Workshop}, vol.~1, pp.~pp--286, 2002.

\bibitem{Drop2016Exact}
P.~Drap and J.~Lef{\`e}vre, ``An exact formula for calculating inverse radial
  lens distortions,'' {\em Sensors}, vol.~16, no.~6, p.~807, 2016.

\bibitem{Abramowitz64}
M.~Abramowitz and I.~A. Stegun, {\em Handbook of Mathematical Functions with
  Formulas, Graphs, and Mathematical Tables}.
\newblock New York: Dover, ninth dover printing, tenth gpo printing~ed., 1964.

\bibitem{tech-report:Montibeller-Perez}
A.~Montibeller and F.~P\'erez-Gonz\'alez, ``Technical report additional
  material for “an adaptive method for camera attribution under complex
  radial distortion corrections”,'' 2022.
\newblock \\ \url{http://dx.doi.org/10.13140/RG.2.2.20038.96323}.

\bibitem{Widrow85}
B.~Widrow and S.~Stearns, {\em Adaptive Signal Processing}.
\newblock Englewood Cliffs, NJ: Prentice-Hall, 1985.

\bibitem{Perez2016Fast}
F.~P{\'e}rez-Gonz{\'a}lez, M.~Masciopinto, I.~Gonz{\'a}lez-Iglesias, and
  P.~Comesa{\~n}a, ``Fast sequential forensic detection of camera
  fingerprint,'' in {\em 2016 IEEE International Conference on Image Processing
  (ICIP)}, pp.~3902--3906, IEEE, 2016.

\bibitem{Goljan2018Blind}
M.~Goljan, ``Blind detection of image rotation and angle estimation,'' {\em
  Electronic Imaging}, vol.~2018, no.~7, pp.~158--1, 2018.

\bibitem{mandelli2019facing}
S.~Mandelli, P.~Bestagini, L.~Verdoliva, and S.~Tubaro, ``Facing device
  attribution problem for stabilized video sequences,'' {\em IEEE Transactions
  on Information Forensics and Security}, vol.~15, pp.~14--27, 2019.

\bibitem{iuliani2021leak}
M.~Iuliani, M.~Fontani, and A.~Piva, ``A leak in prnu based source
  identification—questioning fingerprint uniqueness,'' {\em IEEE Access},
  vol.~9, pp.~52455--52463, 2021.

\end{thebibliography}
\end{document}